\documentclass[10pt,twocolumn,letterpaper]{article}

\usepackage[pagenumbers]{cvpr} 

\usepackage{graphicx}
\usepackage{amsmath}
\usepackage{amssymb}
\usepackage{booktabs}
\usepackage[super]{nth}
\usepackage[page]{appendix}
\usepackage{pifont} 

\usepackage{bm}
\usepackage{colortbl}
\usepackage{scalerel,amssymb}
\usepackage{wasysym}
\makeatletter
\@namedef{ver@everyshi.sty}{}
\makeatother
\usepackage{tikz}
\newcommand*\circled[1]{\tikz[baseline=(char.base)]{
            \node[shape=circle,draw,inner sep=1.pt] (char) {#1};}}

\newcommand{\parsection}[1]{\vspace{0.5mm}\noindent\textbf{#1:}~}

\newcommand*\makeSet[1]{\mathcal{#1}}

\DeclareMathOperator*{\argmin}{arg\,min}

\newcommand{\fgenc}{e_\mathrm{fg}}
\newcommand{\testenc}{e_\mathrm{test}}
\newcommand{\bgenc}{e_\mathrm{bg}}
\definecolor{_yellow}{RGB}{255,217,50}
\definecolor{_darkblue}{RGB}{1,25,147}
\definecolor{_red}{RGB}{238,34,12}
\definecolor{_pink}{RGB}{255,64,255}
\definecolor{_orange}{RGB}{255,147,0}
\definecolor{_blue}{RGB}{4,51,255}
\definecolor{_green}{RGB}{97,216,54}
\definecolor{_purple}{RGB}{122,129,255}
\definecolor{_skyblue}{RGB}{86,193,255}
\definecolor{_coral}{RGB}{255,100,78}
\definecolor{_violet}{RGB}{148,55,250}

\definecolor{_superdimp}{RGB}{254,211,23}
\definecolor{_ours}{RGB}{102,3,255}
\definecolor{_other}{RGB}{125,125,125}
\definecolor{_ours_iounet}{RGB}{252,52,51}

\definecolor{_our_mp}{RGB}{254,204,102}
\definecolor{_mo}{RGB}{230,230,230}

\definecolor{_testenc}{RGB}{254,204,102}
\definecolor{_trainenc}{RGB}{255,255,153}
\definecolor{_enc}{RGB}{254,204,153}
\definecolor{_dec}{RGB}{253,154,102}
\definecolor{_weights}{RGB}{20,157,123}

\definecolor{_trainfeat}{RGB}{20,157,123}
\definecolor{_testfeat}{RGB}{253,153,50}
\definecolor{_tar_loc_feat}{RGB}{12,96,172}
\definecolor{_tar_ext_feat}{RGB}{101,50,153}
\definecolor{_test_emb}{RGB}{252,103,51}
\definecolor{_loc_weights}{RGB}{163,0,4}
\definecolor{_reg_wegihts}{RGB}{253,129,123}

\newcommand{\method}[1]{$\begingroup\color{#1}\CIRCLE\endgroup$}
\newcommand{\weights}[1]{$\begingroup\color{#1}\CIRCLE\endgroup$}
\newcommand{\bbox}[1]{$\begingroup\color{#1}\blacksquare\endgroup$}

\usepackage[pagebackref,breaklinks,colorlinks]{hyperref}

\usepackage[capitalize]{cleveref}
\crefname{section}{Sec.}{Secs.}
\Crefname{section}{Section}{Sections}
\Crefname{table}{Table}{Tables}
\crefname{table}{Tab.}{Tabs.}

\begin{document}

\title{Transforming Model Prediction for Tracking}

\newcommand{\aand}{\hspace{5mm}}
\author{Christoph Mayer \aand Martin Danelljan \aand Goutam Bhat \aand Matthieu Paul \aand Danda Pani Paudel \\ Fisher Yu \aand Luc Van Gool\\
Computer Vision Lab, D-ITET, ETH Z\"urich, Switzerland
}

\maketitle

\begin{abstract}
    Optimization based tracking methods have been widely successful by integrating a target model prediction module, providing effective global reasoning by minimizing an objective function. While this inductive bias integrates valuable domain knowledge, it limits the expressivity of the tracking network. 
    In this work, we therefore propose a tracker architecture employing a Transformer-based model prediction module. Transformers capture global relations with little inductive bias, allowing it to learn the prediction of more powerful target models. We further extend the model predictor to estimate a second set of weights that are applied for accurate bounding box regression. The resulting tracker relies on training and on test frame information in order to predict all weights transductively. We train the proposed tracker end-to-end and validate its performance by conducting comprehensive experiments on multiple tracking datasets. Our tracker sets a new state of the art on three benchmarks, achieving an AUC of 68.5\% on the challenging LaSOT~\cite{Fan_2019_CVPR_Lasot} dataset. The code and trained models are available at \mbox{\url{https://github.com/visionml/pytracking}}
\end{abstract}

\section{Introduction}

Generic visual object tracking is one of the fundamental problems in computer vision. The task involves estimating the state of the target object in every frame of a video sequence, given only the initial target location.
One of the key problems in object tracking is learning to robustly detect the target object, given the scarce annotation.
Among exiting methods,  Discriminative Correlation Filters (DCF) ~\cite{Bhat_2019_ICCV_DIMP,Danelljan_2019_CVPR_ATOM,Danelljan_2017_CVPR_ECO,Henriques_2015_TPAMI_KCF,Bolme_2010_CVPR_MOSSE,Galoogahi_2017_ICCV_BACF,Lukezic_2018_IJCV_CSRDCF,Sun_2018_CVPR_DRT} have achieved much success. These approaches learn a target model to localize the target in each frame, by minimizing a discriminative objective function. 
The target model, often set to a convolutional kernel, provides a compact and generalizable representation of the tracked object, leading to the popularity of DCFs. 

\begin{figure}[t]
\centering
\includegraphics[width=1.0\columnwidth, keepaspectratio]{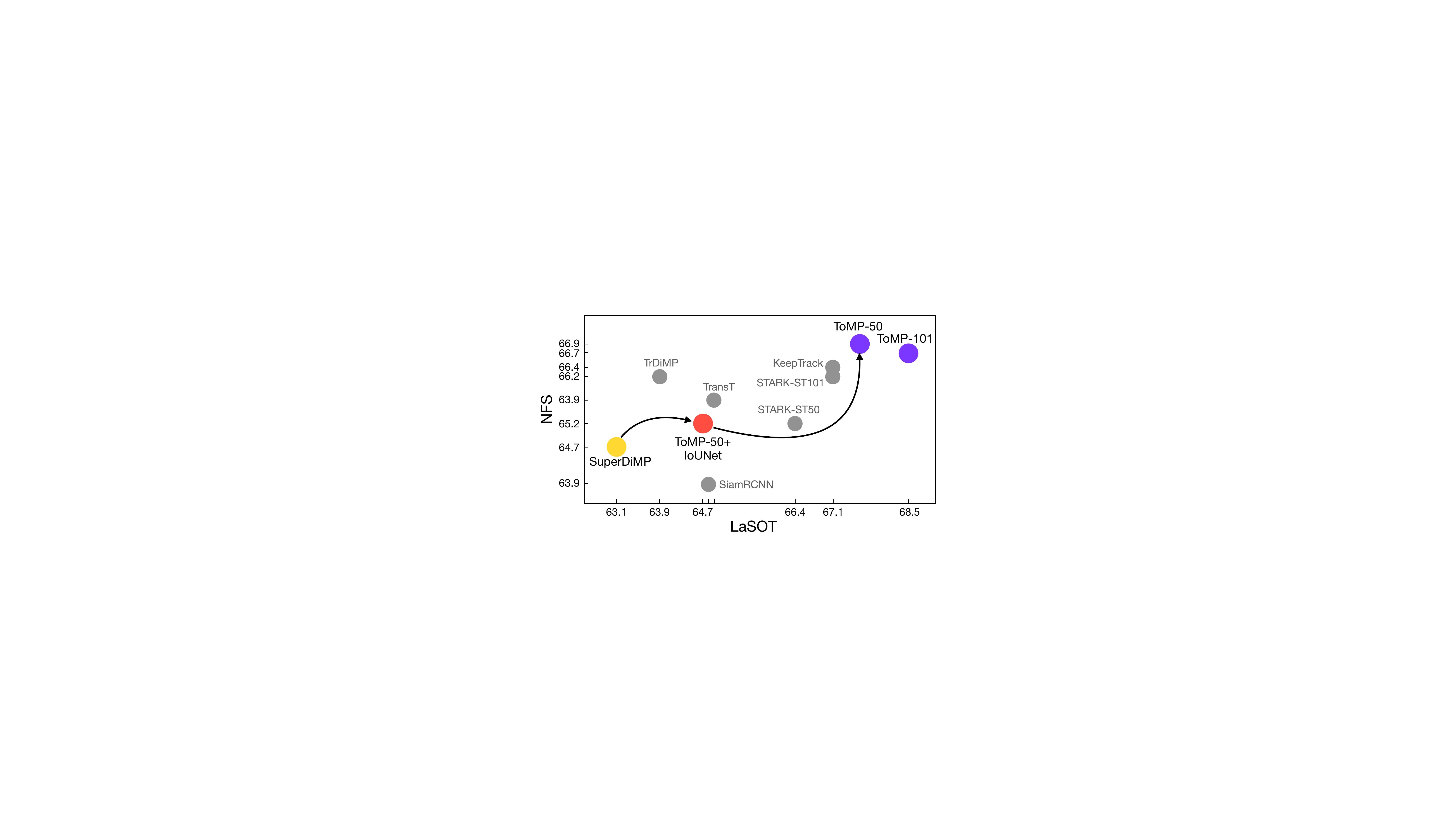}
\caption{Performance improvements when transforming the model optimizer based tracker SuperDiMP~\cite{Danelljan_2019_github_pytracking}~(\method{_superdimp}) step-by-step. First, we replace the model optimizer by a Transformer based model predictor~(\method{_ours_iounet}). Secondly, we replace the probabilistic IoUNet by a new regressor and predict its weights with the same model predictor~(\method{_ours}). The performance (success AUC) is reported on NFS~\cite{Galoogahi_2017_ICCV_NFS} and LaSOT~\cite{Fan_2019_CVPR_Lasot} and compared with recent trackers~(\method{_other}). ToMP-50 and ToMP-101 refer to the different employed backbones ResNet-50~\cite{He_2016_CVPR_Resnet} and ResNet-101~\cite{He_2016_CVPR_Resnet}.
}\label{fig:teaser}
\vspace{-5mm}
\end{figure}

The objective function in DCF integrates both foreground and background knowledge over the previous frames, providing effective global reasoning when learning the model.
However, it also imposes severe inductive bias on the predicted target model. 
Since the target model is obtained by solely minimizing an objective over the previous frames, the model predictor has limited flexibility.
For instance, it cannot integrate any learned priors in the predicted target model. On the other hand, Transformers have also been shown to provide strong global reasoning across multiple frames, thanks to the use of self and cross attention. Consequently, Transformers have been applied to generic object tracking~\cite{Chen_2021_CVPR_TransT,Wang_2021_CVPR_TrDiMP,Yan_2021_ICCV_STARK,Yu_2021_ICCV_HPF} with considerable success. 

In this work, we propose a novel tracking framework that aims at bridging the gap between DCF and Transformer based trackers. Our approach employs a compact target model for localizing the target, as in DCF. The weights of this model are however obtained using a Transformer-based model predictor, allowing us to learn more powerful target models, compared to DCFs.  This is achieved by introducing novel encodings of the target state, allowing the Transformer to effectively utilize this information. We further extend our model predictor to generate weights for a bounding box regressor network, in order to condition its predictions on the current target. Our proposed approach ToMP obtains significant improvement in tracking performance compared to state-of-the-art DCF-based methods, while also outperforming recent Transformer based trackers (see Fig.~\ref{fig:teaser}).

\parsection{Contributions} In summary, our main contributions are the following: \textbf{i)} We propose a novel Transformer-based model prediction module in order to replace traditional optimization based model predictors. \textbf{ii)} We extend the model predictor to estimate a second set of weights that are applied for bounding box regression. \textbf{iii)} We develop two novel encodings that incorporate target location and target extent allowing the Transformer-based model predictor to utilize this information. \textbf{iv)} We propose a parallel two stage tracking procedure at test time to decouple target localization and bounding box regression in order to achieve robust and accurate target detection.
\textbf{v)} We perform a comprehensive set of ablation experiments to assess the contribution of each building block of our tracking pipeline and evaluate it on seven tracking benchmarks. The proposed tracker ToMP sets a new state of the art on three including LaSOT~\cite{Fan_2019_CVPR_Lasot} where it achieves an AUC of $68.5\%$ (see Fig.~\ref{fig:teaser}). In addition we show that our tracker ToMP outperforms other Transformer based trackers for every attribute of LaSOT~\cite{Fan_2019_CVPR_Lasot}.  

\section{Related Work}

\parsection{Discriminative Model Prediction}
DCF based approaches learn a target model to distinguish the target from background by minimizing an objective. For long Fourier-transform based solvers were predominant for DCF based trackers~\cite{Bolme_2010_CVPR_MOSSE,Henriques_2015_TPAMI_KCF,Danelljan_2016_ECCV_CCOT,Lukezic_2018_IJCV_CSRDCF}.
Danelljan~\etal~\cite{Danelljan_2019_CVPR_ATOM} employed a two layer Perceptron as target model and use Conjugate Gradient to solve the optimization problem.
Recently, multiple methods have been introduced that enable end-to-end training by casting the tracking problem into a meta-learning problem~\cite{Wang_2020_CVPR_MAML,Bhat_2019_ICCV_DIMP,Zheng_2020_ECCV_DCFST}. These methods are based on the idea of unrolling the iterative optimization algorithm for a fixed number of iterations and to integrate it in the tracking pipeline to allow end-to-end training. Bhat~\etal~\cite{Bhat_2019_ICCV_DIMP} learn a discriminative feature space and predict the weights of the target model based on the target state in the initial frame and refine the weights with an optimization algorithm. 

\parsection{Transformers for Tracking}
Recently, several trackers have been introduced that use Transformers~\cite{Chen_2021_CVPR_TransT,Yu_2021_ICCV_HPF,Yan_2021_ICCV_STARK,Wang_2021_CVPR_TrDiMP}. Transformers are typically employed to predict discriminative features to localize the target object and regress its bounding box. The training features are processed by the Transformer Encoder whereas the Transformer Decoder fuses training and test features using cross attention layers to compute discriminative features~\cite{Wang_2021_CVPR_TrDiMP,Yu_2021_ICCV_HPF,Chen_2021_CVPR_TransT}.

DTT~\cite{Yu_2021_ICCV_HPF} feeds these features to two networks that predict the location and the bounding box of the target.
In contrast, TransT\cite{Chen_2021_CVPR_TransT} employs a feature fusion network that consists of multiple self and cross attention modules. The fused output features are fed into a target classifier and a bounding box regressor.
TrDiMP~\cite{Wang_2021_CVPR_TrDiMP} adopts the DiMP~\cite{Bhat_2019_ICCV_DIMP} model predictor to produce the model weights given the output features of the Transformer Encoder as training samples. Afterwards, the target model computes the target score map by applying the predicted weights on the output features produced by the Transformer Decoder. TrDiMP adopts the probabilistic IoUNet~\cite{Danelljan_2020_CVPR_PRDIMP} for bounding box regression. Similar to our tracker, TrDiMP encodes target state information but integrates it via two different cross attention modules in the Decoder instead of using two encoding modules in front of the Transformer.

In contrast to the aforementioned Transformer based trackers, STARK~\cite{Yan_2021_ICCV_STARK} adopts the Transformer architecture from DETR~\cite{Carion_2020_ECCV_DETR}. Instead of fusing the training and test features in the Transformer Decoder they are stacked and processed jointly by the full Transformer. A single object-query then produces the Decoder output that is fused with the Transformer Encoder features. These features are then further processed to directly predict the bounding box of the target. In contrast, our tracker employs the same Transformer architecture from DETR~\cite{Carion_2020_ECCV_DETR} but to replace the model optimizer. In the end, our resulting Transformer-based model predictor estimates the weights of two separate models: the target classifier and the bounding box regressor.
\begin{figure*}[t]
\centering
\begin{subfigure}{0.49\linewidth}
    \includegraphics[width=\columnwidth, keepaspectratio]{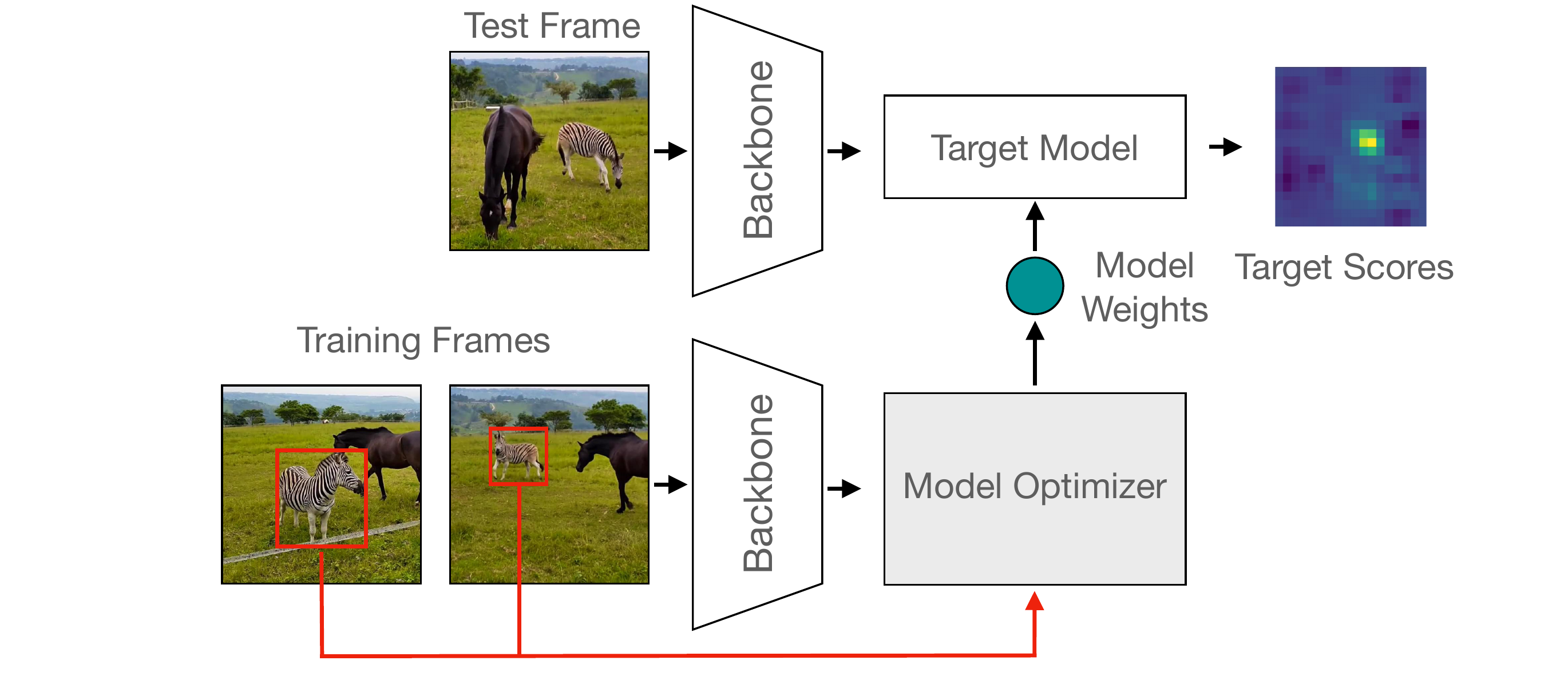}
    \caption{Tracker with optimization based model prediction.}
    \label{fig:overeview-dcf}
  \end{subfigure}
  \hfill
  \begin{subfigure}{0.49\linewidth}
    \includegraphics[width=\columnwidth, keepaspectratio]{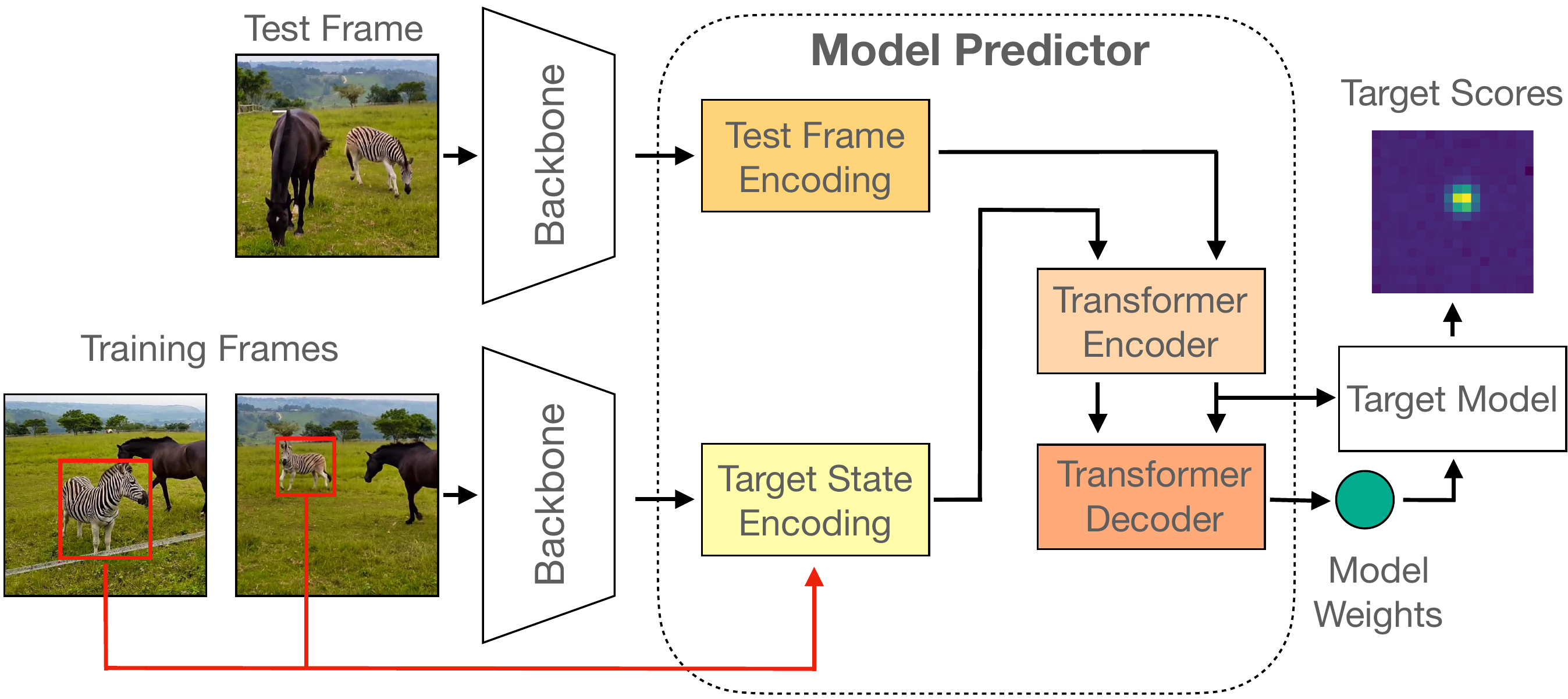}
    \caption{Proposed tracker with Transformer based model prediction.}
    \label{fig:overeview-transformer}
  \end{subfigure}
\caption{Comparison between trackers that employ optimization based model prediction and our Transformer-based model prediction. The model optimizer [\bbox{_mo}] in Fig.~\ref{fig:overeview-dcf}  is replaced by the model predictor in Fig.~\ref{fig:overeview-transformer} that consists of the proposed modules [\bbox{_trainenc},\bbox{_testenc},\bbox{_enc},\bbox{_dec}].
}\label{fig:overview}
\vspace{-0mm}
\end{figure*}

\section{Method}

In this work, we propose a Transformer-based target model prediction network for tracking called ToMP. 
We first revisit existing optimization based model predictors and discuss their limitations in Sec.~\ref{sec:background}. Next, we describe our Transformer-based model prediction approach in Sec.~\ref{sec:trans_model_pred}. We extend this approach to perform joint target classification and bounding box regression in Sec.~\ref{sec:joint_mode_pred}. Finally, we detail our offline training procedure and online tracking pipeline in Sec.~\ref{sec:offline_training} and Sec.~\ref{sec:online_tracking}, respectively.

\subsection{Background}
\label{sec:background}

One of the popular paradigms for visual object tracking is discriminative model prediction based tracking. These approaches, visualized in Fig.~\ref{fig:overeview-dcf}, use a target model to localize the target object in the test frame. The weights (parameters) of this target model are obtained from the model optimizer, using the training frames and their annotation.
While a variety of target models are used in the literature~\cite{Danelljan_2019_CVPR_ATOM,Bhat_2019_ICCV_DIMP,Zheng_2020_ECCV_DCFST,Wang_2020_CVPR_MAML,Lukezic_2018_IJCV_CSRDCF,Sun_2018_CVPR_DRT,Kenan_2019_CVPR_ASRCF}, discriminative trackers share a common base formulation to produce the target model weights.  
This involves solving an optimization problem such that the target model produces the desired target states $y_i\in \makeSet{Y}$ for the training samples $\makeSet{S}_\mathrm{train} \in \{(x_i, y_i)\}_{i=1}^{m}$. Here, $x_i\in\makeSet{X}$ refers to a deep feature map of frame $i$ and $m$ denotes the total number of training frames. The optimization problem reads as follows,
\begin{equation}\label{eq:online-dcf-objective}
    w = \argmin_{\tilde{w}} \sum_{(x,y)\in\makeSet{S}_\mathrm{train}} f(h(\tilde{w}; x), y) + \lambda g(\tilde{w}).
\end{equation}
Here, the objective consists of the residual function $f$ which computes an error between the target model output $h(\tilde{w}; x)$ and the ground truth label $y$. $g(\tilde{w})$ denotes the regularization term weighted by a scalar $\lambda$, while $w$ represents the optimal weights of the target model. Note that the training set $\makeSet{S}_\mathrm{train}$ contains the annotated first frame, as well as the previous tracked frames with the tracker's predictions being used as pseudo-labels. 

Learning the target model by explicitly minimizing the objective of~\eqref{eq:online-dcf-objective} provides a robust target model that can distinguish the target from the previously seen background. However, such a strategy suffers from notable limitations. The optimization based methods compute the target model using only limited information available in previously tracked frames. That is, they cannot integrate learned priors in the target model prediction so as to minimize \emph{future} failures. 
Similarly, these methods typically lack the possibility to utilize the current test frame in a transductive manner when computing the model weights to improve tracking performance.
The optimization based methods also require setting multiple optimizer hyper-parameters, and can overfit/underfit on the training samples.
Another limitation of optimization based trackers is their procedure that produces the discriminative features.
Usually, the features provided to the target model are simply the extracted test features. Instead of reinforced features by using the target state information contained in the training frames. Extracting such enhanced features would allow reliable differentiation between the target and background regions in the test frame. 

\begin{figure*}[t]
\centering
\includegraphics[width=\textwidth, keepaspectratio]{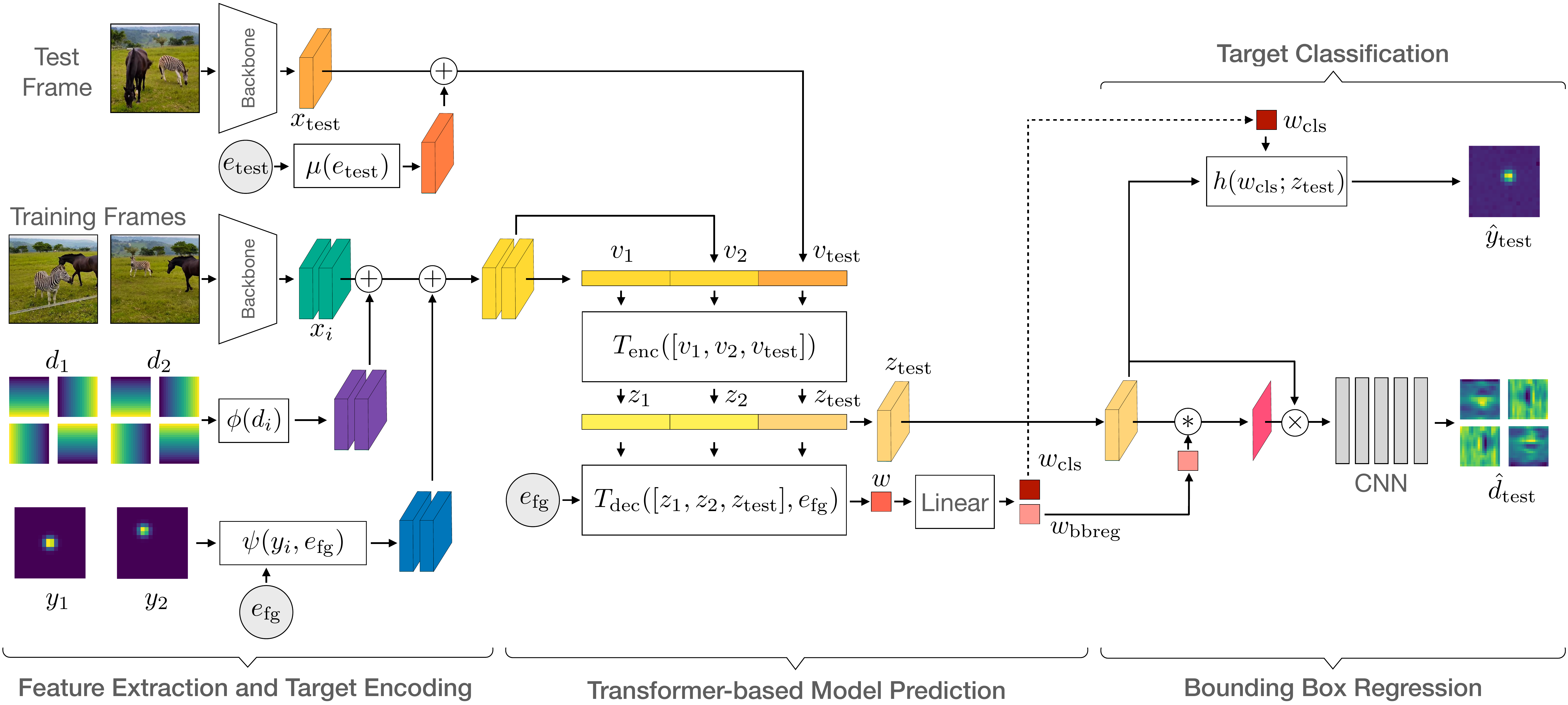}
\vspace{-6mm}
\caption{Overview of the entire ToMP tracking pipeline for joint model prediction. First, the training [\bbox{_trainfeat}] and test [\bbox{_testfeat}] features are extracted using a backbone. Then the target location [\bbox{_tar_loc_feat}] and bounding box [\bbox{_tar_ext_feat}] encodings are added to the training features. For the test frame the test embedding is encoded [\bbox{_test_emb}] and added to the test features. The features are then concatenated and jointly processed by the Transformer-based model predictor that produces the weights used for target classification [\bbox{_loc_weights}] and bounding box regression [\bbox{_reg_wegihts}].
}\label{fig:diagram}
\vspace{-5mm}
\end{figure*}

\subsection{Transformer-based Target Model Prediction}
\label{sec:trans_model_pred}

In order to overcome the aforementioned limitations of optimization based target localization approaches, we propose to replace the model optimizer by a novel target model predictor based on Transformers (see Fig.~\ref{fig:overeview-transformer}).
Instead of explicitly minimizing an objective as stated in~\eqref{eq:online-dcf-objective}, our approach learns to directly predict the target model purely from data by end-to-end training. This allows the model predictor to integrate target specific priors in the predicted model so that it can focus on characteristic features of the target, in addition to the features that allow to differentiate the target from the \emph{seen} background. Furthermore, our model predictor also utilizes the current test frame features, in addition to the previous training features, to predict the target model in a transductive manner. As a result, the model predictor can utilize the current frame information to predict a more suitable target model. Finally, instead of applying the target model on a fixed feature space, defined by the pre-trained feature extractor, our approach can utilize the target information to dynamically construct a more discriminative feature space for every frame.

An overview of the proposed tracker employing the Transformer-based model prediction is shown in Fig.~\ref{fig:overeview-transformer}. Similar to the optimization based trackers, it consists of a test and training branch. We first encode the target state information in the training frames and fuse it with the deep image features [\bbox{_trainenc}]. Similarly, we also add an encoding to the test frame in order to mark it as test frame [\bbox{_testenc}]. The features from both the training and test branches are then jointly processed in the Transformer Encoder [\bbox{_enc}] that produces enhanced features by reasoning globally across frames. Next, the Transformer Decoder [\bbox{_dec}] predicts the target model weights [\weights{_weights}] using the output of the Transformer Encoder. Finally, the predicted target model is applied on the enhanced test frame features to localize the target. Next, we describe the main components in our tracking pipeline.

\parsection{Target Location Encoding}
We propose a target location encoding that allows the model predictor to incorporate the target state information from the training frames, when predicting the target model.
In particular, we use the embedding $\fgenc\in\mathbb{R}^{1\times C}$ that represents \textit{foreground}. Together with 
a Gaussian $y_i\in\mathbb{R}^{H\times W \times 1}$ centered at the target location, we define the target encoding function
\begin{equation}
    \psi(y_i, \fgenc) = y_i\cdot \fgenc,
\end{equation}
where "$\cdot$" denotes point-wise multiplication with broadcasting. Note, that $H_\mathrm{im}=s\cdot H$ and $W_\mathrm{im}=s\cdot W$ correspond to the spatial dimension of the image patch and $s$ to the stride of the backbone network used to extract the deep features $x\in \mathbb{R}^{H\times W \times C}$.
Next, we combine the target encoding and the deep image features $x$ as follows 
\begin{equation}\label{eq:loc_encoding_features}
    v_i = x_i + \psi(y_i, \fgenc).
\end{equation}
This provides us the training frame features $v_i\in \mathbb{R}^{H\times W \times C}$ which contain encoded target state information.
Similarly, we also add a test encoding to identify the features corresponding to the test frame as,
\begin{equation}
    v_\mathrm{test} = x_\mathrm{test} + \mu(\testenc),
\end{equation}
where $\mu(\cdot)$ repeats the token $\testenc$ for each patch of $x_\mathrm{test}$.

\parsection{Transformer Encoder} We aim to predict our target model using the foreground and background information from both the training, as well as the test frames. To achieve this, we use a Transformer Encoder~\cite{Vaswani_2017_NIPS_ATTENTION,Carion_2020_ECCV_DETR} module to first jointly process the features from the training frames and the test frame. The Transformer Encoder serves two purposes in our approach. First, as described later, it computes the features used by the Transformer Decoder module to predict the target model. Secondly, inspired by STARK~\cite{Yan_2021_ICCV_STARK}, our Transformer Encoder also outputs enhanced test frame features, which serve as the input to the target model when localizing the target. 

Given multiple encoded training features $v_i \in \mathbb{R}^{H\times W\times C}$ and an encoded test feature $v_\mathrm{test} \in \mathbb{R}^{H\times W\times C}$, we reshape the features to $\mathbb{R}^{(H\cdot W) \times C}$ and concatenate all $m$ training features $v_i$ and the test feature $v_\mathrm{test}$ along the first dimension. These concatenated features are then processed jointly in a Transformer Encoder 
\begin{equation}
    [z_1, \dots, z_m, z_\mathrm{test}] = T_\mathrm{enc}([v_1,\dots,v_m, v_\mathrm{test}]).
\end{equation}
The Transformer Encoder consists of multi-headed self-attention modules~\cite{Vaswani_2017_NIPS_ATTENTION} that enable it to reason globally across a full frame and even across multiple training and test frames. 
In addition, the encoded target state identifies \textit{foreground} and \textit{background} regions and enables the Transformer to differentiate between both regions.

\parsection{Transformer Decoder} The outputs of the Transformer Encoder ($z_i$ and $z_\mathrm{test}$) are used as inputs for the Transformer Decoder~\cite{Vaswani_2017_NIPS_ATTENTION,Carion_2020_ECCV_DETR} to predict the target model weights  
\begin{equation}
    w = T_\mathrm{dec}([z_1, \dots, z_m, z_\mathrm{test}], \fgenc).
\end{equation}
Note that the inputs $z_i$ and $z_\mathrm{test}$ are obtained by jointly reasoning over the whole training and test samples, allowing us to predict a discriminative target model. We use the same learned \textit{foreground} embedding $\fgenc$ as used for target state encoding as input query of the Transformer Decoder such that the Decoder predicts the target model weights.

\parsection{Target Model}
We use the DCF target model to obtain the target classification scores
\begin{equation}
\label{eq:targetmodel}
    h(w, z_\mathrm{test}) = w \ast z_\mathrm{test}.
\end{equation}
Here, the weights of the convolution filter $w \in \mathbb{R}^{1\times C}$ are predicted by the Transformer Decoder. Note that the target model is applied on the output test features $z_\mathrm{test}$ of the Transformer Encoder. These features are obtained after joint processing of training and test frames, and thus support the target model to reliably localize the target.   

\subsection{Joint Localization and Box Regression}
\label{sec:joint_mode_pred}
In the previous section, we presented our Transformer based architecture for predicting the target model. Although the target model can localize the object center in each frame, a tracker needs to also estimate an accurate bounding box of the target. DCF based trackers typically employ a dedicated bounding box regression network~\cite{Danelljan_2019_CVPR_ATOM} for this task. 
While it is possible to follow a similar strategy, we decide to predict both models jointly since target localization and bounding box regression are related tasks that can benefit from one another. In order to achieve this, we extend our model as follows. First, instead of only using the target center location when generating the target state encoding, we also encode target size information to provide a richer input to our model predictor. Secondly, we extend our model predictor to estimate weights for a bounding box regression network, in addition to the target model weights. The resulting tracking architecture is visualized in Fig.~\ref{fig:diagram}. Next, we describe each of these changes in detail.

\parsection{Target Extent Encoding}
In addition to the extracted deep image features $x_i$ and the target location encoding $\psi(y_i, \fgenc)$, we add another encoding to incorporate information about the bounding box of the target.
In order to encode the bounding box $b_i=\{b^x_i, b^y_i, b^w_i, b^h_i\}$ encompassing the target object in the training frame $i$, we adopt the \textit{ltrb} representation~\cite{Tian_2019_ICCV_FCOS,Fu_2021_CVPR_STMTrack,Xu_2020_AAAI_SiamFCpp,Yu_2021_ICCV_HPF}. First, we map each location $(j^x,j^y)$ on the feature map $x_i$ back to the image domain using $(k^x, k^y) = (\lfloor\frac{s}{2}\rfloor + s\cdot j^x, \lfloor\frac{s}{2}\rfloor + s\cdot j^y)$. Then, we compute the normalized distance of each remapped location to the four sides of the bounding box $b_i$ as follows,
\begin{equation}\label{eq:ltrb}
    \begin{aligned}
        l_i &= (k^x - b^x_i)/W_\mathrm{im}, & r_i&= (k^x - b^x_i - b^w_i)/W_\mathrm{im}, \\
        t_i &= (k^y - b^y_i)/H_\mathrm{im}, & b_i&= (k^y - b^y_i - b^h_i)/H_\mathrm{im},
    \end{aligned}
\end{equation}
where $W_\mathrm{im} = s\cdot W$ and $H_\mathrm{im} = s\cdot H$.
These four sides are used to produce the dense bounding box representation $d = (l, t, r, b)$,  where $d\in \mathbb{R}^{H\times W \times 4}$. In this representation, we encode the bounding box using a Multi-Layer Perceptron (MLP) $\phi$ and thereby increase the number of dimensions from $4$ to $C$ before adding the obtained encoding to Eq.~\eqref{eq:loc_encoding_features} such that
\begin{equation}
    v_i = x_i + \psi(y_i, \fgenc) + \phi(d_i).
\end{equation}
Here, $v_i$ is the resulting feature map which is used as input to the Transformer Encoder, see Fig.~\ref{fig:diagram}.

\parsection{Model Prediction} We extend our architecture to predict weights for the target model, as well as bounding box regression. Concretely, we pass the output $w$ of the Transformer Decoder through a linear layer to obtain the weights for bounding box regression $w_\mathrm{bbreg}$ and target classification $w_\mathrm{cls}$.
The weights $w_\mathrm{cls}$ are then directly used within the target model $h(w_\mathrm{cls};z_\mathrm{test})$ as before. The weights $w_\mathrm{bbreg}$, on the other hand, are used to condition the output test features $z_\mathrm{test}$ of the Transformer Encoder with target information for bounding box regression, as explained next.  

\parsection{Bounding Box Regression}
To make the encoder output features $z_\mathrm{test}$ target aware, we follow Yan~\etal~\cite{Yan_2021_ICCV_STARK} and first compute an attention map $w_\mathrm{bbreg}\ast z_\mathrm{test}$ using the predicted weights $w_\mathrm{bbreg}$. The attention weights are then multiplied point-wise with the test features $z_\mathrm{test}$ before feeding them into a Convolutional Neural Network (CNN). The last layer of the CNN uses an exponential activation function to produce the normalized bounding box prediction in the same \textit{ltrb} representation as described in Eq.~\eqref{eq:ltrb}. In order to obtain the final bounding box estimation, we first extract the center location by applying the $\mathrm{argmax}(\cdot)$ function on the target score map $\hat{y}_\mathrm{test}$ predicted by the target model. 
Next, we query the dense bounding box prediction $\hat{d}_\mathrm{test}$ at the center location of the target object to obtain the bounding box. 
We use two dedicated networks for target localization and bounding box regression in contrast to Yan~\etal~\cite{Yan_2021_ICCV_STARK} that uses one network trying to predict both. This allows us as explained in Sec.~\ref{sec:online_tracking} to decouple target localization from bounding box regression during tracking.

\subsection{Offline Training}\label{sec:offline_training}

In this section, we describe the protocol to train the proposed tracker ToMP. Similar to recent end-to-end trained discriminative trackers~\cite{Bhat_2019_ICCV_DIMP,Danelljan_2020_CVPR_PRDIMP}, we sample multiple training and test frames from a video sequence to form training sub-sequences. In particular, we use two training frames and one test frame. In contrast to recent Transformer based trackers~\cite{Chen_2021_CVPR_TransT,Yan_2021_ICCV_STARK,Yu_2021_ICCV_HPF} but similar to DCF based trackers~\cite{Danelljan_2019_CVPR_ATOM,Bhat_2019_ICCV_DIMP,Danelljan_2020_CVPR_PRDIMP}, we keep the same spatial resolution for training and test frames. We pair each image $I_i$ with the corresponding bounding box $b_i$. We use the target state of the training frames to encode target information and use the bounding box of the test frame only to supervise training by computing two losses based on the predicted bounding boxes and the derived center location of the target in the test frame. 

We employ the target classification loss from DiMP~\cite{Bhat_2019_ICCV_DIMP} that consists of different losses for background and foreground regions.
Further, we employ the generalized Intersection over Union loss~\cite{Rezatofighi_2019_CVPR_GIOU} using the \textit{ltrb} bounding box representation~\cite{Tian_2019_ICCV_FCOS} to supervise bounding box regression
\begin{equation}
    L_\mathrm{tot} = \lambda_\mathrm{cls} L_\mathrm{cls}(\hat{y}, y) + \lambda_\mathrm{giou}L_\mathrm{giou}(\hat{d}, d),
\end{equation}
where $\lambda_\mathrm{cls}$ and $\lambda_\mathrm{giou}$ are scalars weighting the contribution of each loss. 
Note that in contrast to FCOS~\cite{Tian_2019_ICCV_FCOS} and related trackers~\cite{Fu_2021_CVPR_STMTrack} we omit an additional centerness loss since it would be redundant in addition to our classification loss that serves the same purpose. A detailed study examining the impact of centerness is available in the supplementary.

\parsection{Training Details}
We train our tracker on the training splits of the LaSOT~\cite{Fan_2019_CVPR_Lasot}, GOT10k~\cite{Huang_2021_TPAMI_GOT10k}, Trackingnet~\cite{2018_Muller_Trackingnet} and MS-COCO~\cite{Lin_2014_ECCV_COCO} datasets. We sample 40k sub-sequences and train for 300 epochs on two Nvidia Titan RTX GPUs. We use ADAMW~\cite{Loshchilov_2019_ICLR_ADAMW} with a learning rate of 0.0001 that we decay by a factor of 0.2 after 150 and 250 epochs and weight decay of 0.0001. We set $\lambda_\mathrm{cls}=100$ and $\lambda_\mathrm{giou}=1$. We construct a training sub-sequence by randomly sampling two training frames and a test frame from a 200 frame window within a video sequence. We then extract the image patches after randomly translating and scaling the image relative to the target bounding box. Moreover, we use random image flipping and color jittering for data augmentation. We set the spatial resolution of the target scores to $18\times18$ and set the search area scale factor to 5.0. Further training and architecture details are provided in the supplementary Sec.~\ref{sup:sec:details}.

\subsection{Online Tracking}\label{sec:online_tracking}
During tracking, we use the annotated first frame, as well as previously tracked frames as our training set $\makeSet{S}_\mathrm{train}$.
While we always keep the initial frame and its annotation, we include one previously tracked frame and replace it with the most recent frame that achieves a target classifier confidence higher than a threshold. Hence, the training set $\makeSet{S}_\mathrm{train}$ contains at most two frames. 

We observed that incorporating previous tracking results in $\makeSet{S}_\mathrm{train}$ improves the target localization considerably.. However, including predicted bounding box estimations degrades the bounding box regression performance due to inaccurate predictions, see Sec.~\ref{sec:ablation}. Hence, we run the model predictor twice. First, we include intermediate predictions in $\makeSet{S}_\mathrm{train}$ to obtain the classifier weights. In the second pass, we only use the annotated initial frame to predict the bounding box. Note that for efficiency both steps can be performed in parallel in a single forward pass. In particular, we reshape the feature map corresponding to two training and one test frame to a sequence and duplicate it. Then, we stack both in the batch dimension to process them jointly with the model predictor. To only allow attention between the initial frame with ground truth annotation and the test frame when predicting the model for bounding box regression, we make use of the so-called \textit{key\_padding\_mask} that allows to ignore certain keys when computing attention.

\section{Experiments}

We evaluate our proposed tracking architecture ToMP on seven benchmarks. Our approach is based on PyTorch 1.7 and is developed within the PyTracking~\cite{Danelljan_2019_github_pytracking} frame work.
PyTracking is available under the GNU GPL 3.0 license.
On a single Nvidia RTX 2080Ti GPU, ToMP-101 and ToMP-50 achieve 19.6 and 24.8 FPS and use a ResNet-101~\cite{He_2016_CVPR_Resnet} and ResNet-50~\cite{He_2016_CVPR_Resnet} as backbone respectively.

\subsection{Ablation Study}\label{sec:ablation}

We perform a comprehensive analysis of the proposed tracker. First, we analyze the contribution of the different proposed target state encodings and then examine the effect of different inference settings. Finally, we report the performance achieved when replacing the target classifier or the bounding box regressor of SuperDiMP with ours. All ablation experiments in this part use a ResNet-50 as backbone.

\begin{table}[!b]
    \vspace{-4mm}
	\centering
    \newcommand{\best}[1]{\textbf{\textcolor{red}{#1}}}
	\newcommand{\scnd}[1]{\textbf{\textcolor{blue}{#1}}}
	\newcommand{\dist}{\hspace{14pt}}%
	\newcommand{\yes}{\textcolor{black}{\checkmark}}
	\newcommand{\no}{\textcolor{black}{\ding{55}}}
	\resizebox{\columnwidth}{!}{%
        \begin{tabular}{c|c@{\dist}c@{\dist}c@{\dist}|c@{\dist}|c@{\dist}|c@{\dist}c@{\dist}c@{\dist}}
        	\toprule
        	 &$\fgenc$   & $\bgenc$  & $\testenc$ & $\phi(\cdot)$ & $q_\mathrm{dec} =\fgenc$ & LaSOT & NFS  & OTB  \\
        	\midrule
            \circled{1}&\no        & \no       & \no        & \yes          & n.a.                     & 66.0         & 64.8        & 68.2 \\
            \circled{2}&\yes       & \no       & \no        & \yes          & \yes                     & 67.1         & \scnd{66.6} & \scnd{70.0} \\
            \circled{3}&\yes       & \yes      & \no        & \yes          & \yes                     & 67.1         & 66.3        & 69.4 \\  
            \circled{4}&\yes       & \no       & \yes       & \yes          & \yes                     & \best{67.6}  & \best{66.9} & \best{70.1} \\
            \circled{5}&\yes       & \yes      & \yes       & \yes          & \yes                     & \scnd{67.4}  & 66.0        & 69.5 \\
            \midrule
            \circled{6}&\yes       & \no       & \yes       & \yes          & \no                      & 66.0         & 66.2        & 69.9 \\
            \circled{7}&\yes       & \no       & \yes       & \no           & \yes                     & 63.1         & 64.2        & 64.0        \\\bottomrule

        \end{tabular}
	}\vspace{-1mm}%
	\caption{For $\fgenc$, $\bgenc$ and $\testenc$ learning the embedding is denoted by \yes~whereas \no~means setting it to zero. Using the encoding $\phi(\cdot)$ is denoted by \yes~whereas \no~ refers to omitting it. For $q_\mathrm{dec} =\fgenc$ the symbol \yes~means sharing the learned embedding $\fgenc$ for encoding and querying the Decoder wheres \no~ means learning two separate embeddings for both tasks. (Our final model is in the \nth{4} row).
	}\vspace{-1mm}
	\label{tab:encodings}%
\end{table}
\parsection{Target State Encoding}
In order to analyze the effect of the different target state encodings we train different variants of our network and evaluate them on multiple datasets. The first five rows of Tab.~\ref{tab:encodings} correspond to versions with different target location encodings. All other settings are kept the same. In addition to the foreground and test embedding, we include a learned background embedding (instead of setting $\bgenc = 0$) to our analysis as follows: $\psi(y_i, \fgenc, \bgenc) = y_i\cdot \fgenc + (1 - y_i)\cdot \bgenc$. However, Tab.~\ref{tab:encodings} shows (\nth{4} vs. \nth{5} row) that adding such a learned background embedding decreases the tracking performance. We further observe that setting the foreground embedding $\fgenc = 0$ (\nth{1} row) and only relying on the target extent encoding $\phi(\cdot)$ still achieves high tracking performance but clearly lacks behind all other versions that include the foreground embedding.
We conclude that using only the foreground encoding $\fgenc$ and the test encoding $\testenc$ leads to the best performance (\nth{4} row).

In the second part of Tab.~\ref{tab:encodings} we choose the best settings for the target location encoding and remove either the target extent encoding $\phi(\cdot)$ or decouple the Transformer Decoder query from the foreground embedding $\fgenc$. We observe that using a separate query (\nth{6} row) decreases the overall performance. Similarly, we notice that incorporating target extent information via the proposed encoding is crucial. Otherwise, the performance drops significantly (\nth{7} row).

\begin{table}[!t]
	\centering
    \newcommand{\best}[1]{\textbf{\textcolor{red}{#1}}}
	\newcommand{\scnd}[1]{\textbf{\textcolor{blue}{#1}}}
	\newcommand{\dist}{\hspace{10pt}}%
	\newcommand{\yes}{\textcolor{black}{\checkmark}}
	\newcommand{\no}{\textcolor{black}{\ding{55}}}
	\resizebox{\columnwidth}{!}{%
        \begin{tabular}{c@{\dist}c@{\dist}c@{\dist}|c@{\dist}c@{\dist}c@{\dist}}
        	\toprule
        	Number of        &                 & Decoder                           &             &             &\\
        	Decoder queries  & Linear Layer    & query $q_\mathrm{dec}$            & LaSOT       & NFS         & OTB         \\
        	\midrule
        	1                & \yes            & $q_\mathrm{dec} = \fgenc$         & \best{67.6} & \best{66.9} & \best{70.1}        \\
            2                & \no             & $q_\mathrm{dec} \neq \fgenc$      & 63.7        & 62.8        & 67.9 \\
\bottomrule
        \end{tabular}
	}\vspace{-2mm}%
	\caption{Analysis of different model predictor architectures and its impact on the tracking performance in terms of success AUC.
	}\vspace{-2mm}
	\label{tab:ablation-model-predictor}%
\end{table}
\parsection{Model Predictor}
Since our model predictor estimates two different model weights, it seems natural to use two different Transformer queries: one to produce the target model weights and the other to obtain the bounding box regressor weights. However, this involves decoupling the query from the foreground embedding $\fgenc$ and the experiments in Tab.~\ref{tab:ablation-model-predictor} show a significant performance drop for this case.


\begin{table}[!t]
	\centering
    \newcommand{\best}[1]{\textbf{\textcolor{red}{#1}}}
	\newcommand{\scnd}[1]{\textbf{\textcolor{blue}{#1}}}
	\newcommand{\dist}{\hspace{14pt}}%
	\newcommand{\yes}{\textcolor{black}{\checkmark}}
	\newcommand{\no}{\textcolor{black}{\ding{55}}}
	\resizebox{\columnwidth}{!}{%
        \begin{tabular}{c@{\dist}c@{\dist}|c@{\dist}c@{\dist}c@{\dist}}
        	\toprule
        	Two Stage         &  Previous          &             &             &             \\
    	    Model Prediction  & Tracking Results   & LaSOT       & NFS         & OTB         \\
        	\midrule
        	n.a.              & \no                & 65.7        & 65.3        & 67.8        \\
            \yes              & \yes               & \best{67.6} & \best{66.9} & \best{70.1} \\
            \no               & \yes               & 62.0        & 64.8        & 62.8        \\
\bottomrule
        \end{tabular}
	}\vspace{-2mm}%
	\caption{Analysis of different inference settings an of their impact on the tracking performance in terms of success AUC.
	}\vspace{-2mm}
	\label{tab:ablation-inference}%
\end{table}
\parsection{Inference Settings}
During online tracking, we use the initial frame and its annotation as training frames. In addition, we include the most recent frame and its target prediction if the classifier confidence is above a certain threshold.
Tab.~\ref{tab:ablation-inference} shows that including previous tracking results leads to higher tracking performance than using only the initial frame.
Disabling the described two stage model prediction approach and predicting the weights of the target model and bounding box regressor at once decreases the tracking performance drastically (-5.6 AUC on LaSOT). The reason is the sensitivity of the bounding box predictor to inaccurate predicted boxes that are encoded and used for training.

\begin{table}[t]
	\centering
	\newcommand{\best}[1]{\textbf{\textcolor{red}{#1}}}
	\newcommand{\scnd}[1]{\textbf{\textcolor{blue}{#1}}}
	\newcommand{\dist}{\hspace{14pt}}%
	\newcommand{\yes}{\textcolor{black}{\checkmark}}
	\resizebox{1.\columnwidth}{!}{%
        \begin{tabular}{c@{\dist}c@{\dist}|c@{\dist}c@{\dist}c@{\dist}c@{\dist}c@{\dist}c@{\dist}}
        	\toprule
        	        Model     & Bounding Box &             &             &             & LaSOT\\
        	        Predictor & Regressor    & LaSOT       & NFS         & UAV         & ExtSub\\
        	\midrule
                    DiMP~\cite{Bhat_2019_ICCV_DIMP}      & Prob. IoUNet~\cite{Danelljan_2020_CVPR_PRDIMP} & 63.1        & 64.8        & \scnd{67.7} & 43.7\\
                    \textbf{ToMP}                        & Prob. IoUNet~\cite{Danelljan_2020_CVPR_PRDIMP} & \scnd{64.7} & \scnd{65.2} & 65.0        & \scnd{45.2} \\  
                    \textbf{ToMP}                        & \textbf{ToMP}                                  & \best{67.6} & \best{66.9} & \best{69.0} & \best{45.4}
        \\\bottomrule
        \end{tabular}
	}\vspace{-2mm}
	\caption{Impact of replacing DiMP~\cite{Bhat_2019_ICCV_DIMP} and the probabilistic IoUNet~\cite{Danelljan_2020_CVPR_PRDIMP} with ToMP for localization and box regression.
	}\vspace{-6mm}
	\label{tab:transforming_model_prediction}%
\end{table}

\parsection{Transforming Model Prediction Step-by-Step}
Our model predictor can estimate model weights for the target model and bounding box regressor. In this part, we will transform an optimization based tracker step-by-step to assess the impact of each transformation step. 
Tab.~\ref{tab:transforming_model_prediction} shows that replacing the model optimizer in SuperDiMP (\nth{1} row) with our proposed model predictor to only predict the target model
(\nth{2} row) outperforms SuperDiMP on three out of four datasets. Our tracker ToMP that jointly predicts model weights for target localization and bounding box regression (\nth{3} row) achieves the best performance on all four datasets.
We conclude that predicting the weights of the target model improves the performance and likewise predicting the weights of the bounding box regressor. Note that we report the average over five runs for all trackers based on the probabilistic IoUNet due to its stochasticity.

\subsection{Comparison to the State of the Art}\label{subsec:experiments}
We compare our tracker ToMP on seven tracking benchmarks. 
The same settings and parameters are used for all datasets.  We recompute the metrics of all trackers using the raw predictions if available or otherwise report the results given in the respective paper.

\begin{table}[!b]
	\centering
	\vspace{-3mm}
	\newcommand{\best}[1]{\textbf{\textcolor{red}{#1}}}
	\newcommand{\scnd}[1]{\textbf{\textcolor{blue}{#1}}}
	\newcommand{\opt}[1]{\textbf{\textcolor{violet}{#1}}}
	\newcommand{\fast}[1]{\textbf{\textcolor{orange}{#1}}}
	\newcommand{\dist}{\hspace{3pt}}%
	\resizebox{1.00\linewidth}{!}{%
        \begin{tabular}{l@{\dist}c@{\dist}c@{\dist}c@{\dist}c@{\dist}c@{\dist}c@{\dist}c@{\dist}c@{\dist}c@{\dist}c@{\dist}c@{\dist}c@{\dist}c@{\dist}c@{\dist}c@{\dist}c@{\dist}c@{\dist}c@{\dist}c@{\dist}c@{\dist}c@{\dist}c@{\dist}c@{\dist}c@{\dist}c@{\dist}c@{\dist}c@{\dist}c@{\dist}c@{\dist}}
        	\toprule
        	           & \textbf{ToMP} & \textbf{ToMP} & STARK & Keep  & STARK & Alpha &        & Siam  & Tr   & Super &      &STM   &     & Pr    \\
        	           & \textbf{101} & \textbf{50}   & ST101 & Track & ST50  &Refine & TransT & R-CNN & DiMP & DiMP  & SAOT &Track & DTT & DiMP  \\
        	           &              &               &\cite{Yan_2021_ICCV_STARK} & \cite{Mayer_2021_ICCV_KeepTrack} &\cite{Yan_2021_ICCV_STARK} & \cite{Yan_2021_CVPR_AlphaRefine} & \cite{Chen_2021_CVPR_TransT} & \cite{Voigtlaender_2020_CVPR_SiamRCNN} & \cite{Wang_2021_CVPR_TrDiMP} & \cite{Danelljan_2019_github_pytracking} & \cite{Zhou_2021_ICCV_SAOT} &\cite{Fu_2021_CVPR_STMTrack} & \cite{Yu_2021_ICCV_HPF} & \cite{Danelljan_2020_CVPR_PRDIMP}  \\
        	\midrule
        	
        	Precision      & \best{73.5} & \scnd{72.2} & \scnd{72.2} & 70.2 & 71.2 & 68.0 & 69.0 & 68.4 & 66.3 & 65.3 & -    & 63.3 & -    & 60.8 \\
        	Norm. Prec     & \best{79.2} & \scnd{78.0} & 76.9        & 77.2 & 76.3 & 73.2 & 73.8 & 72.2 & 73.0 & 72.2 & 70.8 & 69.3 & -    & 68.8 \\
        	Success (AUC)  & \best{68.5} & \scnd{67.6} & 67.1        & 67.1 & 66.4 & 65.3 & 64.9 & 64.8 & 63.9 & 63.1 & 61.6 & 60.6 & 60.1 & 59.8 \\\bottomrule
        \end{tabular}
	}\vspace{-3mm}
	\caption{Comparison on the LaSOT~\cite{Fan_2019_CVPR_Lasot} test set ordered by AUC.
	}\vspace{-1mm}
	\label{tab:lasot}%
\end{table}
\begin{figure}[b]
\centering
\includegraphics[width=\columnwidth, keepaspectratio]{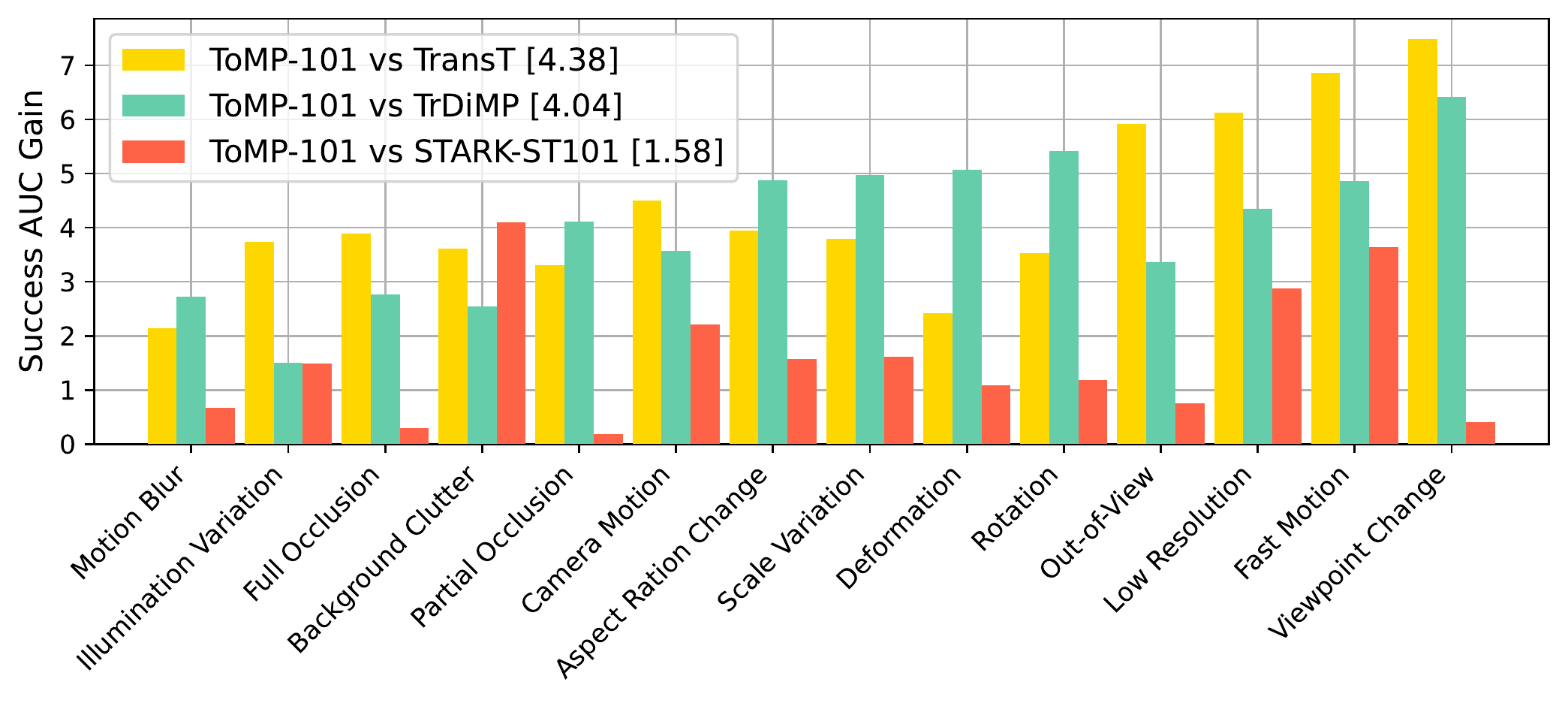}
\vspace{-5mm}

\caption{Per attribute analysis on LaSOT~\cite{Fan_2019_CVPR_Lasot} between ToMP and recent Transformer based trackers. The bar heights correspond to the gain of our tracker and the legend shows the average gain.}\label{fig:comarison_to_transformers}\vspace{-0mm}
\end{figure}

\begin{figure}[t]
\centering
\begin{subfigure}{0.49\columnwidth}
    \includegraphics[width=\columnwidth, keepaspectratio]{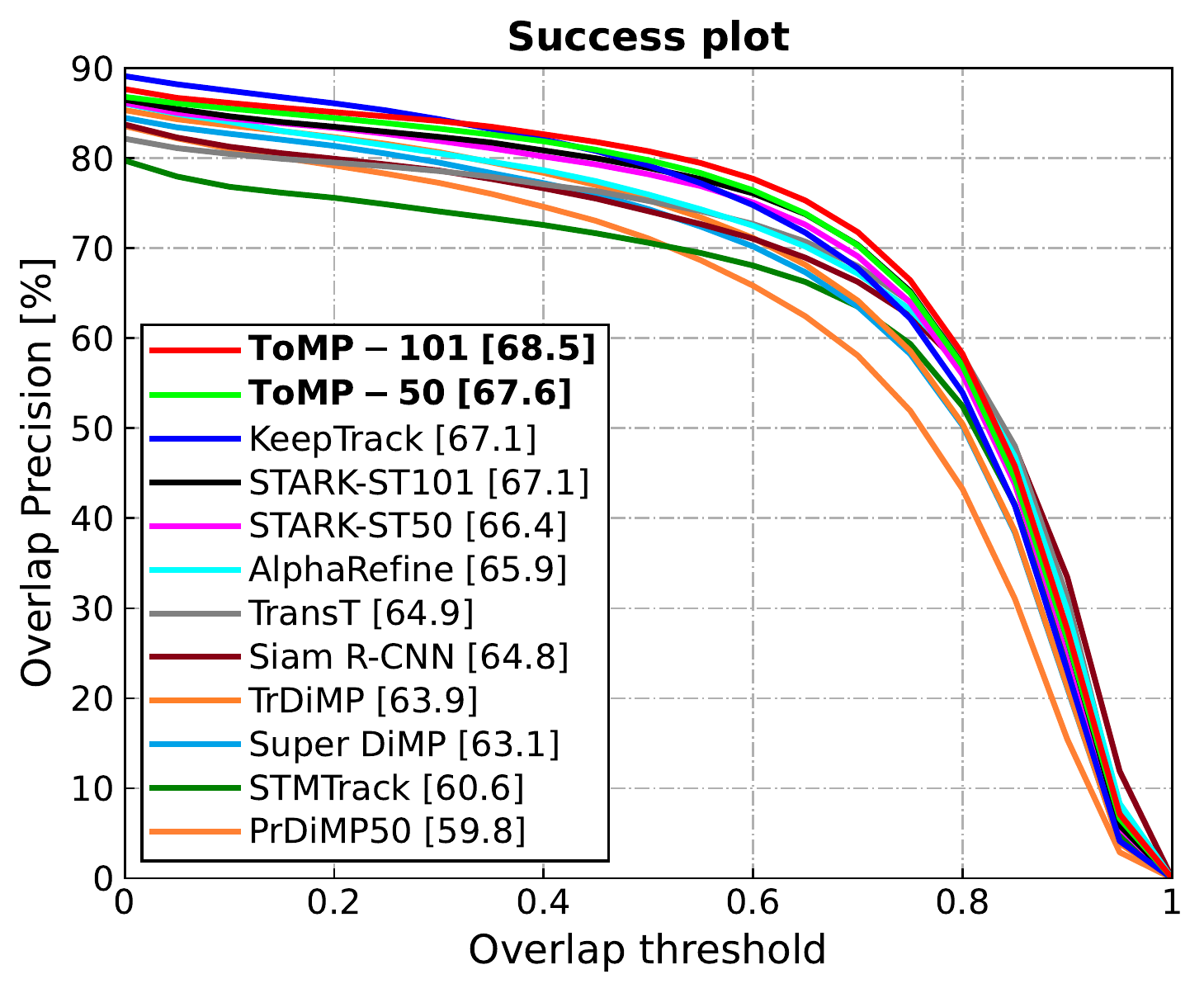}
    \caption{LaSOT~\cite{Fan_2019_CVPR_Lasot}}
    \label{fig:lasot}
  \end{subfigure}
  \hfill
  \begin{subfigure}{0.49\columnwidth}
    \includegraphics[width=\columnwidth, keepaspectratio]{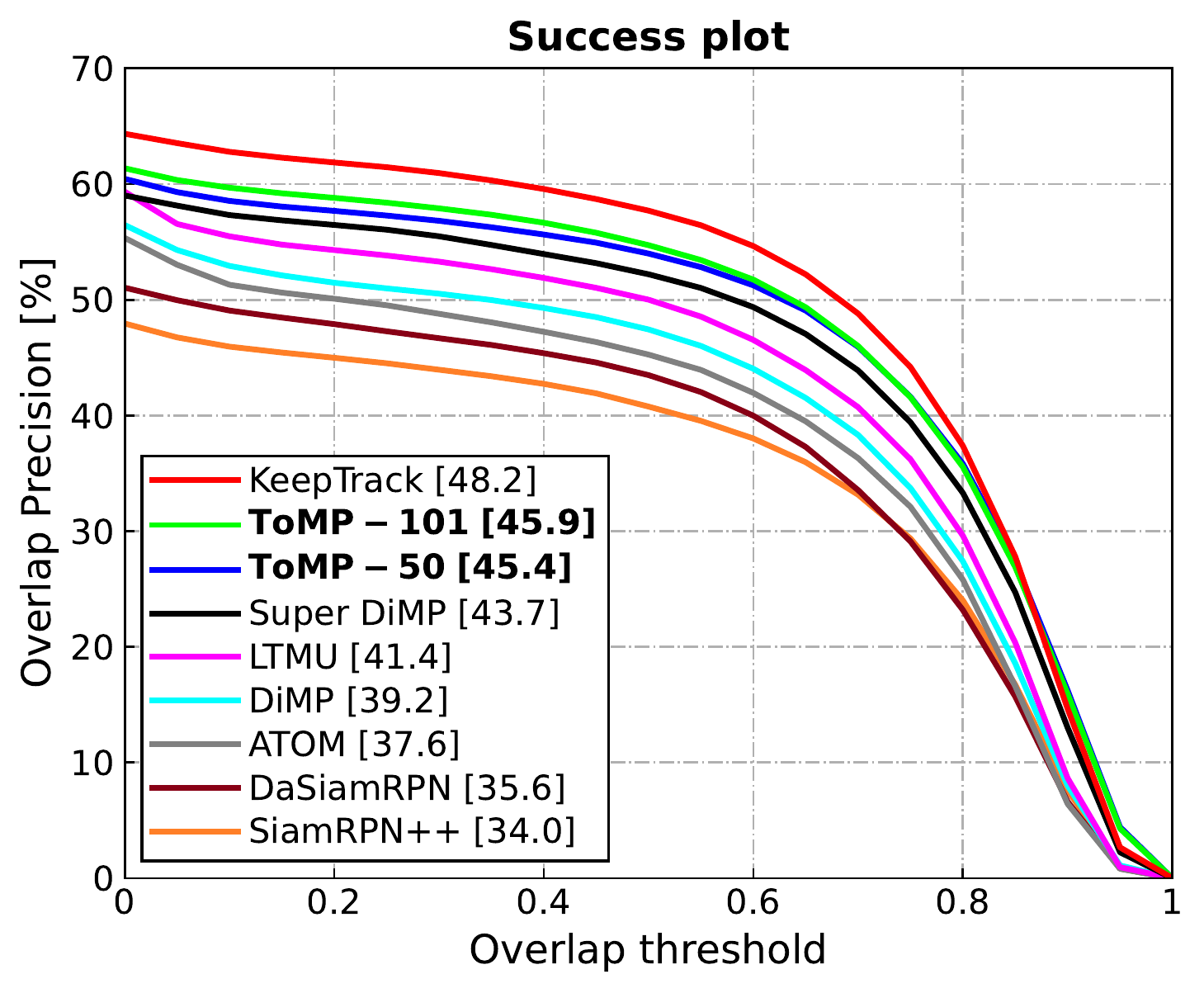}
    \caption{LaSOTExtSub~\cite{Fan_2020_IJCV_Lasot_ext}}
    \label{fig:lasot_ext_sub}
  \end{subfigure}
\caption{Success plots, showing $\text{OP}_T$, on LaSOT~\cite{Fan_2019_CVPR_Lasot} and LaSOTExtSub~\cite{Fan_2020_IJCV_Lasot_ext} and AUC is reported in the legend.}\label{fig:success}
\vspace{-3mm}
\end{figure}

\parsection{LaSOT~\cite{Fan_2019_CVPR_Lasot}}
First, we compare ToMP on the large-scale LaSOT dataset (280 test sequences with 2500 frames on average). The success plot in Fig.~\ref{fig:lasot} shows the overlap precision $\text{OP}_T$ as a function of the threshold $T$. 
Trackers are ranked \wrt their \emph{area-under-the-curve} (AUC) score, shown in the legend. Tab.~\ref{tab:lasot} shows more results including precision and normalized precision for each tracker. Both versions of ToMP with different backbones outperform the recent trackers STARK~\cite{Yan_2021_ICCV_STARK}, TransT~\cite{Chen_2021_CVPR_TransT}, TrDiMP~\cite{Wang_2021_CVPR_TrDiMP} and DTT~\cite{Yu_2021_ICCV_HPF} in AUC and sets a new state-of-the-art result.
Note that even ToMP with ResNet-50 outperforms STARK-ST101 with ResNet-101 (67.6 vs 67.1). Fig.~\ref{fig:comarison_to_transformers} shows the success AUC gain of ToMP compared to recent Transformer based trackers for different attributes annotated in LaSOT~\cite{Fan_2019_CVPR_Lasot}. We want to highlight that ToMP outperforms TransT~\cite{Chen_2021_CVPR_TransT} and TrDiMP~\cite{Wang_2021_CVPR_TrDiMP} on each attribute by more than one percent point. Similarly, ToMP achieves higher performance than STARK-ST101 for every attribute. It achieves the highest gain over STARK for \textit{Background Clutter}, showing the disadvantage of using small templates
instead of training frames with a large field of view that allow not only to leverage target, but also background information.

\parsection{LaSOTExtSub~\cite{Fan_2020_IJCV_Lasot_ext}}
This dataset is an extension of LaSOT. It only contains test sequences assigned to 15 new classes with 10 videos each. The sequences contain 2500 frames on average showing challenging tracking scenarios of small, fast moving objects with distractors present. Fig.~\ref{fig:lasot_ext_sub} shows the success plot where the results of most trackers are obtained from ~\cite{Fan_2020_IJCV_Lasot_ext}, \eg, DaSiamRPN~\cite{Zhu_2018_ECCV_DaSiamRPN}, SiamRPN++~\cite{Li_2019_CVPR_SiamRPN++}, ATOM~\cite{Danelljan_2019_CVPR_ATOM}, DiMP~\cite{Bhat_2019_ICCV_DIMP} and LTMU~\cite{Dai_2020_CVPR_LTMU}. ToMP exceeds the performance of all trackers except KeepTrack~\cite{Mayer_2021_ICCV_KeepTrack} that employs explicit distractor matching between frames. In particular, we outperform SuperDiMP~\cite{Danelljan_2019_github_pytracking} that uses a model optimizer ($+2.2\%$).

\parsection{TrackingNet~\cite{2018_Muller_Trackingnet}}
We evaluate ToMP on the large-scale TrackingNet dataset that contains 511 test sequences without publicly available ground-truth. An online evaluation server is used to obtain the tracking metrics shown in Tab.~\ref{tab:trackingnet} by submitting the raw tracking results. Both versions of ToMP achieve competitive results close to the current state of the art. In particular, ToMP-101 achieves the second best performance in terms of AUC behind STARK~\cite{Yan_2021_ICCV_STARK}, outperforming other Transformer based trackers such as TransT~\cite{Chen_2021_CVPR_TransT} and TrDiMP~\cite{Wang_2021_CVPR_TrDiMP}.

\parsection{UAV123~\cite{Mueller_2016_ECCV_UAV123}}
The UAV dataset consists of 123 test videos that contains small objects, target occlusion, and distractors. Tab.~\ref{tab:nfs_uav_otb} shows the achieved results in terms of success AUC. Again, ToMP achieves competitive results compared to the current state of the art achieved by KeepTrack~\cite{Mayer_2021_ICCV_KeepTrack}.

\begin{table}[!t]
	\centering
	\newcommand{\best}[1]{\textbf{\textcolor{red}{#1}}}
	\newcommand{\scnd}[1]{\textbf{\textcolor{blue}{#1}}}
	\newcommand{\dist}{\hspace{3pt}}%
	\resizebox{1.00\linewidth}{!}{%
        \begin{tabular}{l@{\dist}c@{\dist}c@{\dist}c@{\dist}c@{\dist}c@{\dist}c@{\dist}c@{\dist}c@{\dist}c@{\dist}c@{\dist}c@{\dist}c@{\dist}c@{\dist}c@{\dist}c@{\dist}c@{\dist}c@{\dist}c@{\dist}c@{\dist}c@{\dist}c@{\dist}c@{\dist}c@{\dist}c@{\dist}c@{\dist}c@{\dist}c@{\dist}c@{\dist}c@{\dist}}
        	\toprule
        	           & \textbf{ToMP} & \textbf{ToMP} & STARK &        & STARK & Siam  & Alpha & STM   &     & Tr   & Keep  & Super & Pr   & Siam \\
        	           & \textbf{101}  & \textbf{50}   & ST101 & TransT & ST50  & R-CNN &Refine & Track & DTT & DiMP & Track & DiMP  & DiMP & FC++ \\
        	           &               &               & \cite{Yan_2021_ICCV_STARK} &\cite{Chen_2021_CVPR_TransT} & \cite{Yan_2021_ICCV_STARK} & \cite{Voigtlaender_2020_CVPR_SiamRCNN} & \cite{Yan_2021_CVPR_AlphaRefine} & \cite{Fu_2021_CVPR_STMTrack} &\cite{Yu_2021_ICCV_HPF} & \cite{Wang_2021_CVPR_TrDiMP} & \cite{Mayer_2021_ICCV_KeepTrack} & \cite{Danelljan_2019_github_pytracking} & \cite{Danelljan_2020_CVPR_PRDIMP} & \cite{Xu_2020_AAAI_SiamFCpp}  \\
        	\midrule
        	
        	Precision      & 78.9        & 78.6 & -           & \best{80.3} & -    & 80.0 & 78.3 & 76.7 & 78.9 & 73.1 & 73.8 & 73.3& 70.4 & 70.5  \\
        	Norm. Prec     & 86.4        & 86.2 & \best{86.9} & \scnd{86.7} & 86.1 & 85.4 & 85.6 & 85.1 & 85.0 & 83.3 & 83.5 & 83.5& 81.6 & 80.0  \\
        	Success (AUC)  & \scnd{81.5} & 81.2 & \best{82.0} & 81.4        & 81.3 & 81.2 & 80.5 & 80.3 & 79.6 & 78.4 & 78.1 & 78.1& 75.8 & 75.4  \\\bottomrule
        \end{tabular}
	}\vspace{-3mm}
	\caption{Comparison on the TrackingNet~\cite{2018_Muller_Trackingnet} test set.
	}\vspace{-3mm}
	\label{tab:trackingnet}%
\end{table}
\begin{table}[!t]
	\centering
	\newcommand{\best}[1]{\textbf{\textcolor{red}{#1}}}
	\newcommand{\scnd}[1]{\textbf{\textcolor{blue}{#1}}}
	\newcommand{\opt}[1]{\textbf{\textcolor{violet}{#1}}}
	\newcommand{\fast}[1]{\textbf{\textcolor{orange}{#1}}}
	\newcommand{\dist}{\hspace{3pt}}%
	\resizebox{1.00\linewidth}{!}{%
        \begin{tabular}{l@{\dist}c@{\dist}c@{\dist}c@{\dist}c@{\dist}c@{\dist}c@{\dist}c@{\dist}c@{\dist}c@{\dist}c@{\dist}c@{\dist}c@{\dist}c@{\dist}c@{\dist}c@{\dist}c@{\dist}c@{\dist}c@{\dist}c@{\dist}c@{\dist}c@{\dist}c@{\dist}c@{\dist}c@{\dist}c@{\dist}c@{\dist}c@{\dist}c@{\dist}c@{\dist}}
        	\toprule
        	        & \textbf{ToMP}& \textbf{ToMP} & Keep        &             & STARK &        &            & STARK     & Super  & Pr    & STM        & Siam & Siam  &      &      \\
        	        & \textbf{101} & \textbf{50}   & Track       & CRACT       & ST101 & TrDiMP & TransT     & ST50      & DiMP   & DiMP  & Track      & AttN & R-CNN & KYS  & DiMP \\
        	        &              &               & \cite{Mayer_2021_ICCV_KeepTrack} & \cite{Fan_2020_arxiv_CRACT} & \cite{Yan_2021_ICCV_STARK} & \cite{Wang_2021_CVPR_TrDiMP} & \cite{Chen_2021_CVPR_TransT} & \cite{Yan_2021_ICCV_STARK} & \cite{Danelljan_2019_github_pytracking} & \cite{Danelljan_2020_CVPR_PRDIMP} & \cite{Fu_2021_CVPR_STMTrack} & \cite{Yu_2020_CVPR_SiamAttN} & \cite{Voigtlaender_2020_CVPR_SiamRCNN} & \cite{Bhat_2020_ECCV_KYS} & \cite{Bhat_2019_ICCV_DIMP}\\          
        	\midrule
        	UAV123  & 66.9         & 69.0          & \best{69.7} & 66.4        & 68.2  & 67.5   & \scnd{69.1}& \scnd{69.1}& 67.7   & 68.0 & 64.7        & 65.0 & 64.9  & --   & 65.3 \\
        	OTB-100 & 70.1         & 70.1          & 70.9        & \best{72.6} & 68.1  & 71.1   & 69.4       & 68.5       & 70.1   & 69.6 & \scnd{71.9} & 71.2 & 70.1  & 69.5 & 68.4 \\
        	NFS     & \scnd{66.7}  & \best{66.9}   & 66.4        & 62.5        & 66.2  & 66.2   & 65.7       & 65.2       & 64.8   & 63.5 & --          & --   & 63.9  & 63.5 & 62.0 \\
            \bottomrule
        \end{tabular}
	}\vspace{-3mm}
	\caption{Comparison with the state of the art on the OTB-100~\cite{WU_2015_TPAMI_OTB}, NFS~\cite{Galoogahi_2017_ICCV_NFS} and UAV123~\cite{Mueller_2016_ECCV_UAV123} datasets in terms of AUC score. 
	}
	\label{tab:nfs_uav_otb}\vspace{-5mm}%
\end{table}

\parsection{OTB-100~\cite{WU_2015_TPAMI_OTB}}
We also report results on the OTB-100 dataset that contains 100 short sequences. Multiple trackers achieve results above 70\% AUC. Among them are both versions of ToMP, see Tab.~\ref{tab:nfs_uav_otb}. ToMP achieve the same performance as SuperDiMP~\cite{Danelljan_2019_github_pytracking} but slightly higher results than TransT~\cite{Chen_2021_CVPR_TransT} and slightly lower than TrDiMP~\cite{Wang_2021_CVPR_TrDiMP}. 

\parsection{NFS~\cite{Galoogahi_2017_ICCV_NFS}}
We compete on the NFS dataset (30FPS version) containing 100 test videos. It contains fast motions and challenging sequences with distractors. Both versions of ToMP exceed the performance of the current best method KeepTrack~\cite{Mayer_2021_ICCV_KeepTrack} by $+0.5\%$ and $+0.3\%$, see Tab.~\ref{tab:nfs_uav_otb}.

\begin{table}[!t]
	\centering
	\vspace{-0mm}
	\newcommand{\best}[1]{\textbf{\textcolor{red}{#1}}}
	\newcommand{\scnd}[1]{\textbf{\textcolor{blue}{#1}}}
	\newcommand{\dist}{\hspace{5pt}}%
	\resizebox{1.00\linewidth}{!}{%
        \begin{tabular}{l@{\dist}c@{\dist}c@{\dist}c@{\dist}c@{\dist}c@{\dist}c@{\dist}c@{\dist}c@{\dist}c@{\dist}c@{\dist}c@{\dist}c@{\dist}c@{\dist}c@{\dist}c@{\dist}c@{\dist}c@{\dist}c@{\dist}c@{\dist}c@{\dist}c@{\dist}c@{\dist}c@{\dist}c@{\dist}c@{\dist}c@{\dist}c@{\dist}c@{\dist}c@{\dist}}
        	\toprule
        	           & \textbf{ToMP} & \textbf{ToMP} & STARK        & Super & STARK         &       &       &       &       &       \\
        	           & \textbf{101}  & \textbf{50}   & ST50         & DiMP  & ST101         & DPMT  & TRAT  & UPDT  & DiMP  & ATOM  \\
        	           & & & \cite{Yan_2021_ICCV_STARK} & \cite{Danelljan_2019_github_pytracking,Kristan_2020_ECCVW_VOT2020} & \cite{Yan_2021_ICCV_STARK} & \cite{Kristan_2020_ECCVW_VOT2020} & \cite{Kristan_2020_ECCVW_VOT2020} & \cite{Bhat_2018_ECCV_UPDT,Kristan_2020_ECCVW_VOT2020} & \cite{Bhat_2019_ICCV_DIMP,Kristan_2020_ECCVW_VOT2020} & \cite{Danelljan_2019_CVPR_ATOM,Kristan_2020_ECCVW_VOT2020} \\          
        	\midrule
        	Accuracy   & 0.453         & 0.453         & 0.478.        & 0.477 & \scnd{0.481} & \best{0.492} & 0.464 & 0.465 & 0.457 & 0.462 \\
        	Robustness & \best{0.814}  & 0.789         & \scnd{0.799}  & 0.728 & 0.775        & 0.745        & 0.744 & 0.755 & 0.734 & 0.734 \\
        	EAO        & \best{0.309}  & 0.297         & \scnd{0.308}  & 0.305 & 0.303        & 0.303        & 0.280 & 0.278 & 0.274 & 0.271 \\\bottomrule

        \end{tabular}
	}\vspace{-3mm}
	\caption{Comparison to the state of the art of bounding box only methods on VOT2020ST~\cite{Kristan_2020_ECCVW_VOT2020} in terms of EAO score.
	}\vspace{-5mm}
	\label{tab:vot2020}%
\end{table}
\parsection{VOT2020~\cite{Kristan_2020_ECCVW_VOT2020}}
Finally, we evaluate on the 2020 edition of the Visual Object Tracking short-term challenge. We compare with the top methods in the challenge~\cite{Kristan_2020_ECCVW_VOT2020}, as well as more recent methods. The dataset contains 60 videos annotated with segmentation masks. Since ToMP produces bounding boxes we only compare with trackers that produce the bounding boxes as well. The trackers are evaluated following the multi-start protocol and are ranked according to the EAO metric that is based on tracking accuracy and robustness, defined using IoU overlap and failure rate respectively. The results in Tab.~\ref{tab:vot2020} show that ToMP-101 achieves the best overall performance, with the highest robustness and competitive accuracy compared to previous methods.

\subsection{Limitations}

Transformer Encoders consist of self-attention layers that compute similarity matrices between multiple training and test frame features and thus lead to a large memory footprint that impacts training and inference run-time. Thus, in future work this limitation should be addressed by evaluating alternatives such as~\cite{Kitaev_2020_ICLR_Reformer,Shen_2021_WACV_EfficientAttention,Katharopoulos_2020_ICML_LinearAttention} aiming at decreasing the memory burden.
Another limiting factor of ToMP arises from challenging tracking sequences. In particular, distractors present while the target is occluded is a typical failure scenario of ToMP, since it is lacking explicit distractor handling as in KeepTrack~\cite{Mayer_2021_ICCV_KeepTrack}. 

\section{Conclusion}

We propose a novel tracking architecture employing a Transformer-based model predictor. The model predictor estimates the weights of the compact DCF target model to localize the target in the test frame. In addition, the predictor produces a second set of weights used for precise bounding box regression. To achieve this, we develop two new modules that encode target location and its bounding box in the training features.  We conduct comprehensive experimental validation and analysis of ToMP on several challenging datasets, and set a new state of the art on three.

{
\noindent\textbf{Acknowledgments:}
This work was partly supported by the ETH Z\"urich Fund (OK), Siemens Smart Infrastructure, the ETH Future Computing Laboratory (EFCL) financed by a gift from Huawei Technologies, an Amazon AWS grant, and an Nvidia hardware grant.}

{\small
\bibliographystyle{ieee_fullname}
\bibliography{egbib}
}
\clearpage
\begin{appendices}
    In this supplementary material, we first provide details about training, model architecture and inference in Sec.~\ref{sup:sec:details}. Further, we report visual results such as a comparison to state-of-the-art trackers, a comparison of different model predictors and failure cases of our tracker in Sec.~\ref{sup:sec:vis-results}. Afterwards, we provide more detailed results of the experiments shown
in the main paper in Sec.~\ref{sup:sec:exps}.

\begin{figure}[b]
\centering
\includegraphics[width=1.0\columnwidth, keepaspectratio]{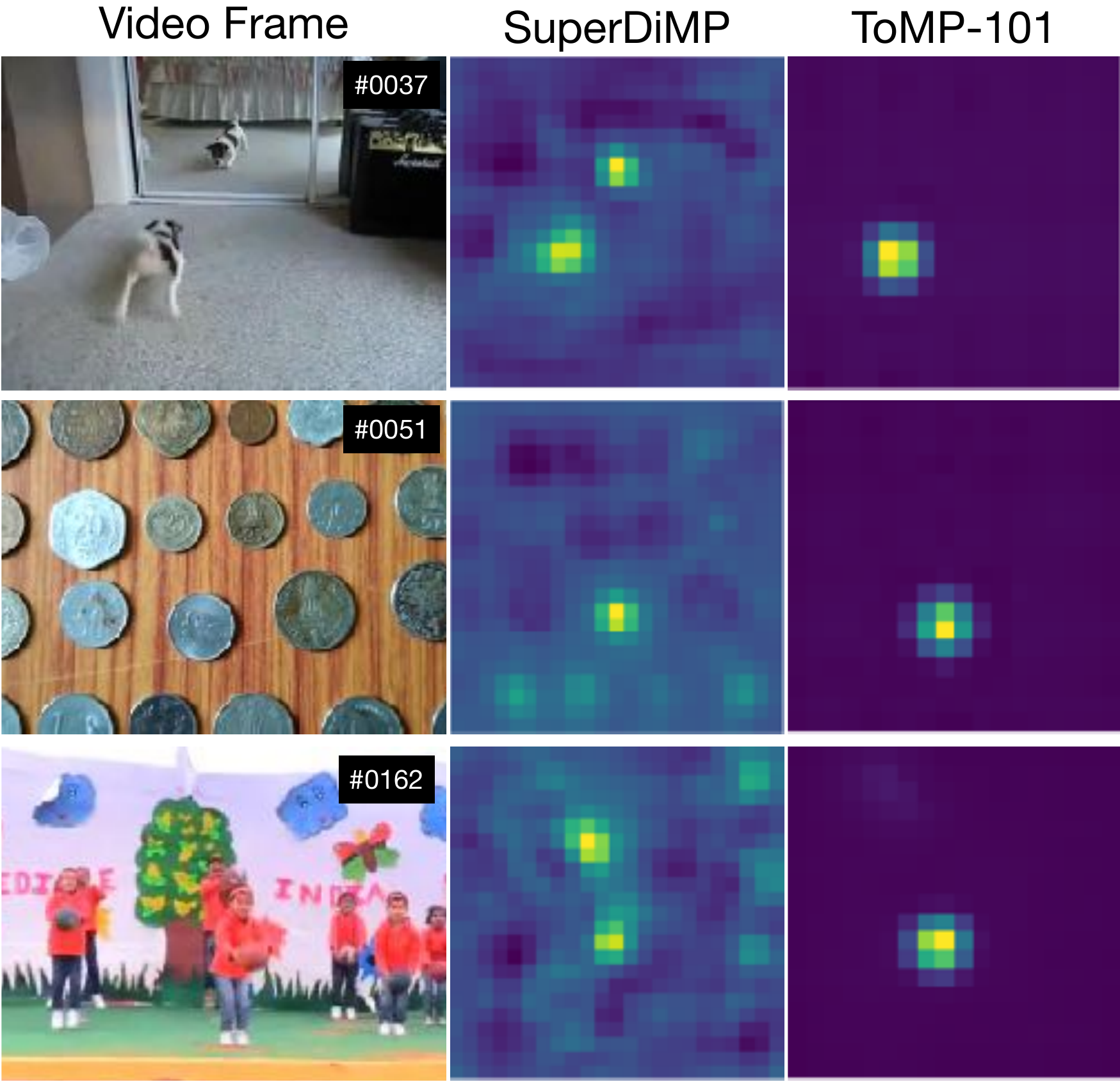}
\caption{Visual comparison of the target score maps resulting from different model predictors. 
}\label{sup:fig:score_maps}
\end{figure}

\section{Training, Architecture and Inference}\label{sup:sec:details}
First, we provide additional details about the training followed by a detailed description of the architectures employed and finally we provide further inference details.

\begin{table}[b]
	\centering
    \newcommand{\best}[1]{\textbf{\textcolor{red}{#1}}}
	\newcommand{\scnd}[1]{\textbf{\textcolor{blue}{#1}}}
	\newcommand{\dist}{\hspace{8pt}}%
	\newcommand{\yes}{\textcolor{black}{\checkmark}}
	\newcommand{\no}{\textcolor{black}{\ding{55}}}
	\resizebox{\columnwidth}{!}{%
        \begin{tabular}{c@{\dist}c@{\dist}c@{\dist}|c@{\dist}c@{\dist}c@{\dist}}
        	\toprule
        	Two Stage         &  Previous          & Confidence       &             &             &             \\
    	    Model Prediction  & Tracking Results   & Threshold $\eta$ & LaSOT       & NFS         & OTB         \\
        	\midrule
        	\yes              & \yes               & 0.85             & 67.3        & 66.9        & \best{70.3} \\
            \yes              & \yes               & 0.90             & \best{67.6} & \best{66.9} & 70.1 \\
            \yes              & \yes               & 0.95             & 67.4        & 66.0        & 69.8 \\
            \bottomrule
        \end{tabular}
	}
	\vspace{-2pt}
	\caption{Analysis of different inference settings an of their impact on the tracking performance in terms of AUC of the success curve.
	}
	\label{sup:tab:ablation-inference}%
	\vspace{-5pt}
\end{table}
\begin{table}[b]
    \vspace{-0pt}%
    \centering%
    \newcommand{\best}[1]{\textbf{\textcolor{red}{#1}}}%
	\newcommand{\scnd}[1]{\textbf{\textcolor{blue}{#1}}}%
	\newcommand{\dist}{\hspace{10pt}}%
    \resizebox{1.\columnwidth}{!}{%
    \begin{tabular}{l@{\dist}|c@{\dist}c@{\dist}c@{\dist}c@{\dist}c@{\dist}|c@{\dist}}
    \toprule
    training frames & NFS & OTB & UAV & LaSOT & LaSOTExtSub & Speed [FPS] \\
    \midrule
    1 initial             & 65.3        & 67.8        & 68.7        & 65.7        & 43.7         & \best{26.2} \\
    1 initial + 1 recent  & 66.9        & 70.1        & 69.0        & \scnd{67.6} & \best{45.4}  & \scnd{24.8} \\
    \midrule
    2 initial + 1 recent  & \best{67.6} & \scnd{70.5} & 67.2        & \best{68.0} & \best{45.4}  & 20.5 \\
    1 initial + 2 recent  & 66.7        & \best{70.8} & \best{69.4} & \scnd{67.6} & 44.4         & 21.8 \\
    1 initial + 3 recent  & 66.8        & \scnd{70.5} & \scnd{69.2} & \scnd{67.6} & 44.2         & 17.6 \\
    1 initial + 4 recent  & \scnd{67.2} & 70.1        & 68.2        & 67.3        & \scnd{44.7}  & 13.2 \\
    1 initial + 5 recent  & 66.8        & 70.1        & 69.1        & 67.2        & 43.9         & 11.3 \\
    \bottomrule
    \end{tabular}}
    \vspace{-2pt}
    \caption{Comparison of different number of training samples in success AUC.}\label{sup:tab:ablation_num_mem_samples}
    \vspace{-5pt}
\end{table}
\begin{table}[b]
    \vspace{0pt}%
    \centering%
    \newcommand{\best}[1]{\textbf{\textcolor{red}{#1}}}%
	\newcommand{\scnd}[1]{\textbf{\textcolor{blue}{#1}}}%
	\newcommand{\no}{\textcolor{black}{\ding{55}}}
    \newcommand{\yes}{\textcolor{black}{\checkmark}}
	\newcommand{\dist}{\hspace{20pt}}%
    \resizebox{1.\columnwidth}{!}{%
    \begin{tabular}{l@{\dist}|c|c@{\dist}c@{\dist}c@{\dist}c@{\dist}c@{\dist}}
    \toprule
                                      & $L_\mathrm{centerness}$ & NFS & OTB & UAV & LaSOT & LaSOTExtSub \\
    \midrule
    Classification                    &\no & \best{66.9} & \best{70.1} & \best{69.0} & \scnd{67.6} & \scnd{45.4} \\
    \midrule
    Classification                    & \yes & \scnd{65.8} & \scnd{69.2} & 67.3        & \best{67.9} & \best{45.5} \\
    Centerness                        & \yes & 62.7        & 66.3        & 67.4        & 64.4        & 41.3        \\
    Classification $\cdot$ Centerness & \yes & 63.7        & 67.8        & \scnd{68.7} & 65.8        & 45.3        \\
    \bottomrule
    \end{tabular}}
    \vspace{-2pt}
    \caption{Impact of centerness scores on training and inference.}\label{sup:tab:centerness}
    \vspace{-5pt}
\end{table}

\begin{figure*}[t!]
\centering
\includegraphics[width=\textwidth, keepaspectratio]{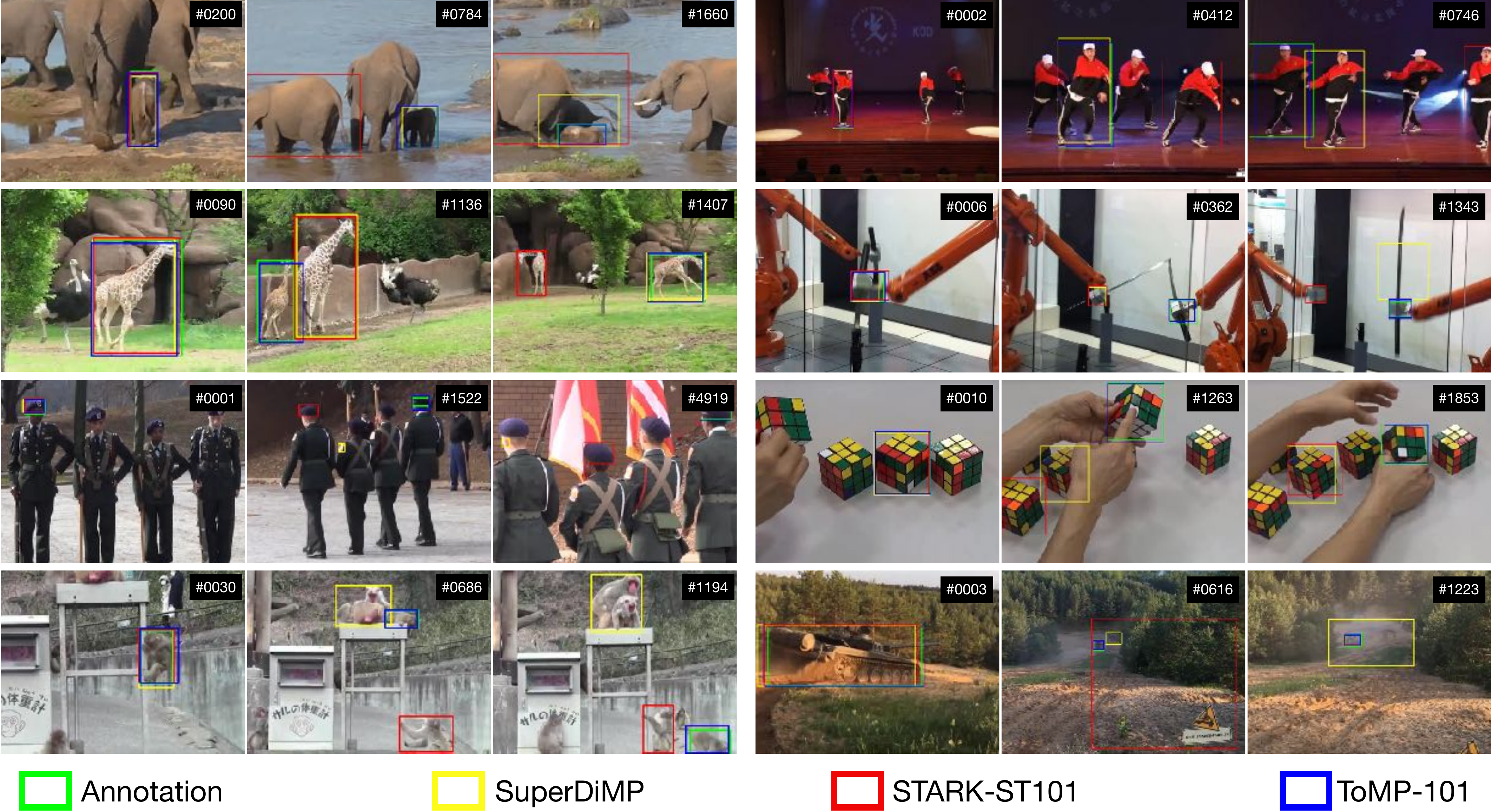}
\caption{Visual comparison of different trackers (ToMP-101, SuperDiMP~\cite{Danelljan_2019_github_pytracking} and STARK-ST101~\cite{Yan_2021_ICCV_STARK}) on different LaSOT~\cite{Fan_2019_CVPR_Lasot} sequences.
}\label{sup:fig:visual-results}
\end{figure*}

\subsection{Training and Architecture Details}\label{sup:sec:train-details}

For training we produce the target states $y$ by using a Gaussian with standard deviation $1/4$ relative to the base target size and by settting $\tau = 0.05$ to differentiate between foreground and background regions in the corresponding classification loss $l_\mathrm{cls}$ adopted from DiMP~\cite{Bhat_2019_ICCV_DIMP}.
For the model predictor we extract features with a stride of 16 from the third block of the ResNet that we use as backbone. We initialize the backbone with the official weights obtained by training the backbone on ImageNet~\cite{Jia_2009_CVPR_ImageNet} and freeze the batch norm statistics during training. Since we use a channel dimension of 256 for the Transformer and the ResNet features have 1024 channels we employ an single convolutional layer to decrease the number of channels before feeding the features into the Transformer Encoder. The Transformer Encoder consists of layers containing multi-headed self attention and a feed-forward network. We use eight heads and a hidden dimension of 2048 for the feed-forward network. Furthermore, we use Dropout with probability 0.1 and layer normalization. The Transformer settings are adopted from DETR~\cite{Carion_2020_ECCV_DETR}.
The predicted target model weights for classification and bounding box regression consist of a single $1\times 1$ filter with 256 channels. The bounding box regression CNN consists of four convolution-instance-normalization-ReLU layers and a final convolution layer, followed by an exponential activation. The MLP for target extent encoding $\phi$ consists of three layers ($4 \rightarrow 64 \rightarrow 256 \rightarrow 256$) where each layer consists of a linear projection, batch normalization and ReLU activation except the last that only consist of a linear projection. The region-encoding tokens $\fgenc$ and $\testenc$ are 256 dimensional learnable embeddings.

\subsection{Inference Details}\label{sup:sec:inf-details}

In order to decide whether a previous tracking result should be used for training of not we use the maximal value of the target score map produced by the target model. In particular, we select the sample if its confidence value is above a certain threshold $\eta$. Tab.~\ref{sup:tab:ablation-inference} shows that the chosen threshold of 0.9 leads to high performance on LaSOT~\cite{Fan_2019_CVPR_Lasot}, NFS~\cite{Galoogahi_2017_ICCV_NFS} and OTB-100~\cite{WU_2015_TPAMI_OTB}.
Furthermore, we follow SuperDiMP~\cite{Danelljan_2019_github_pytracking} and enter in the \textit{target not found} state if the maximal value of the target score map is bellow 0.25. Moreover, we use the same spatial resolution of the target scores of 18 × 18 and the same search area scale factor of 5.0 during inference and training.

Furthermore, we study the effect of using more than two training frames stored in the sample memory. Instead of using only one initial and one recent training frame to predict the network weights we test the impact of increasing the number of recent training frames and of using multiple initial training frames. We increase the number of initial training frames with ground truth bounding box annotations using an augmentation (vertical flipping and random translation). Tab.~\ref{sup:tab:ablation_num_mem_samples} shows the results for different combinations of multiple initial and recent training frames. Note, that we use the same network weights for all experiments trained with one initial and one recent recent frame in all cases. We observer that using more training frames can improve the tracking performance but decreases the run-time. Furthermore, we observe that the tracker greatly benefits from including at least one recent frame for training.

\begin{figure}[b]
\centering
\includegraphics[width=\columnwidth, keepaspectratio]{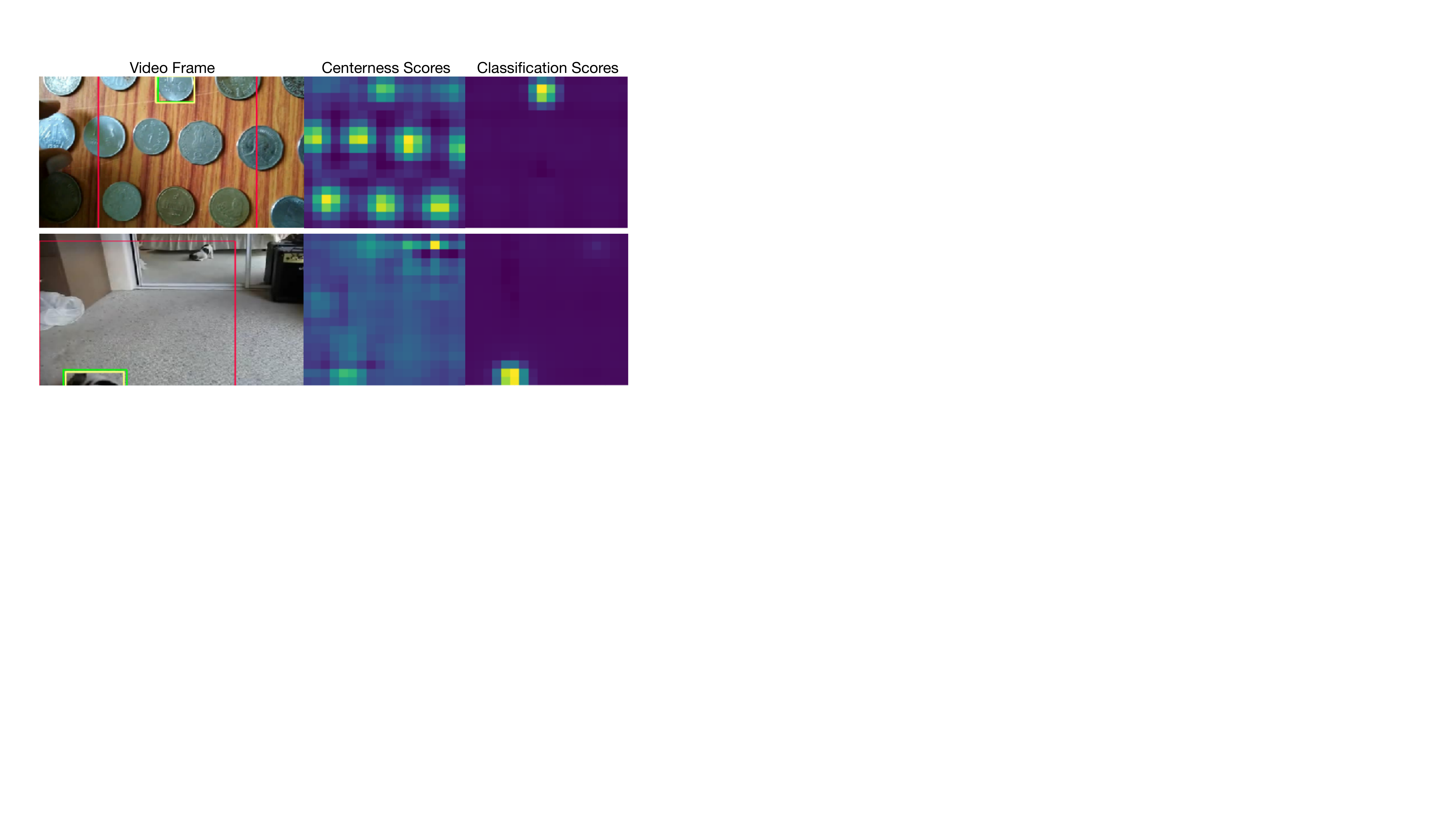}
\caption{Visual Comparison between centerness and classification scores.
}\label{sup:fig:centerness}
\end{figure}

\begin{figure}[b]
\centering
\includegraphics[width=\columnwidth, keepaspectratio]{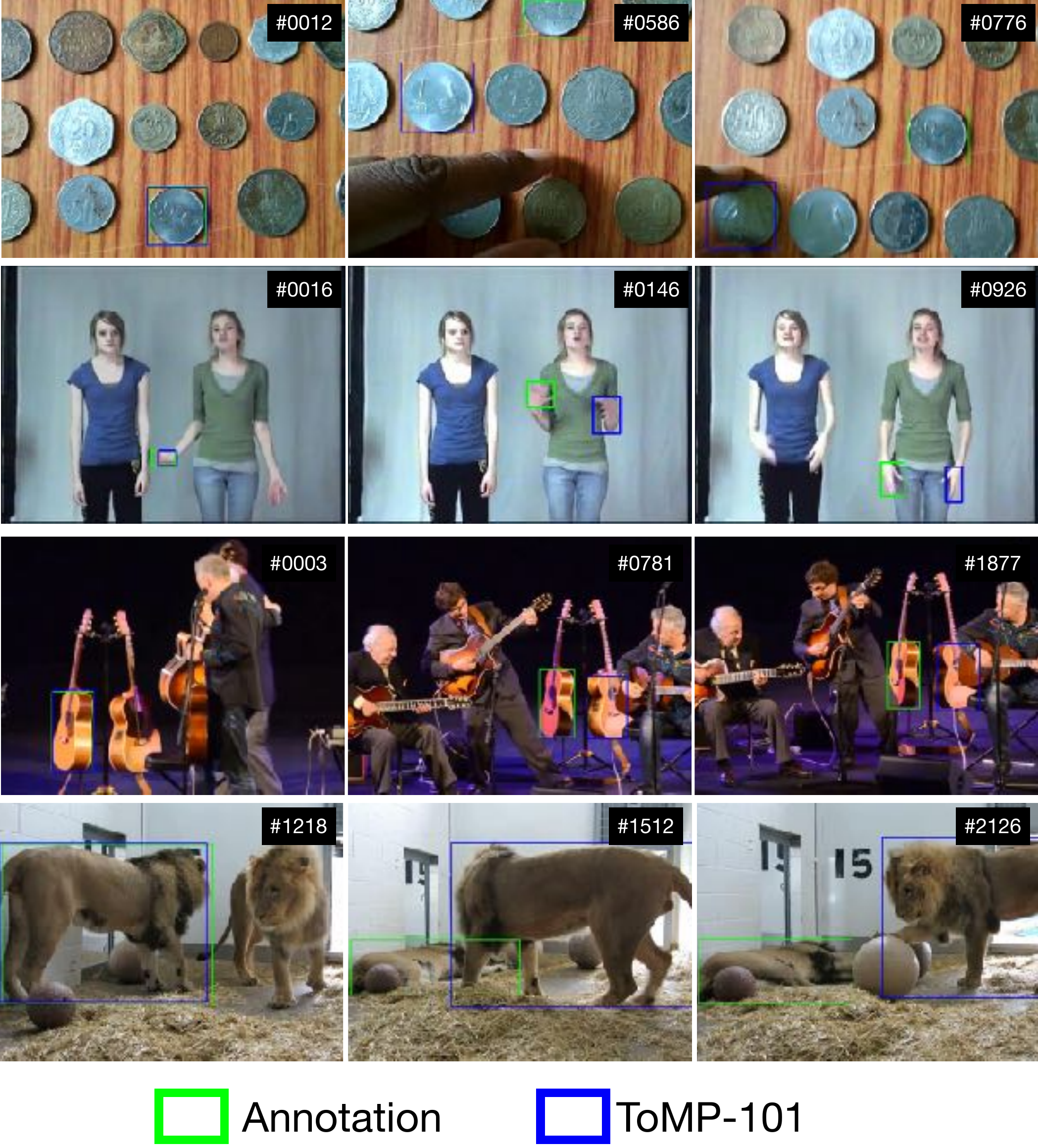}
\caption{Visualization of failure cases of our tracker.
}\label{sup:fig:failure_cases}
\end{figure}

\begin{figure*}[t]
\centering
  \begin{subfigure}{0.33\linewidth}
    \centering
    \includegraphics[width=\linewidth, keepaspectratio]{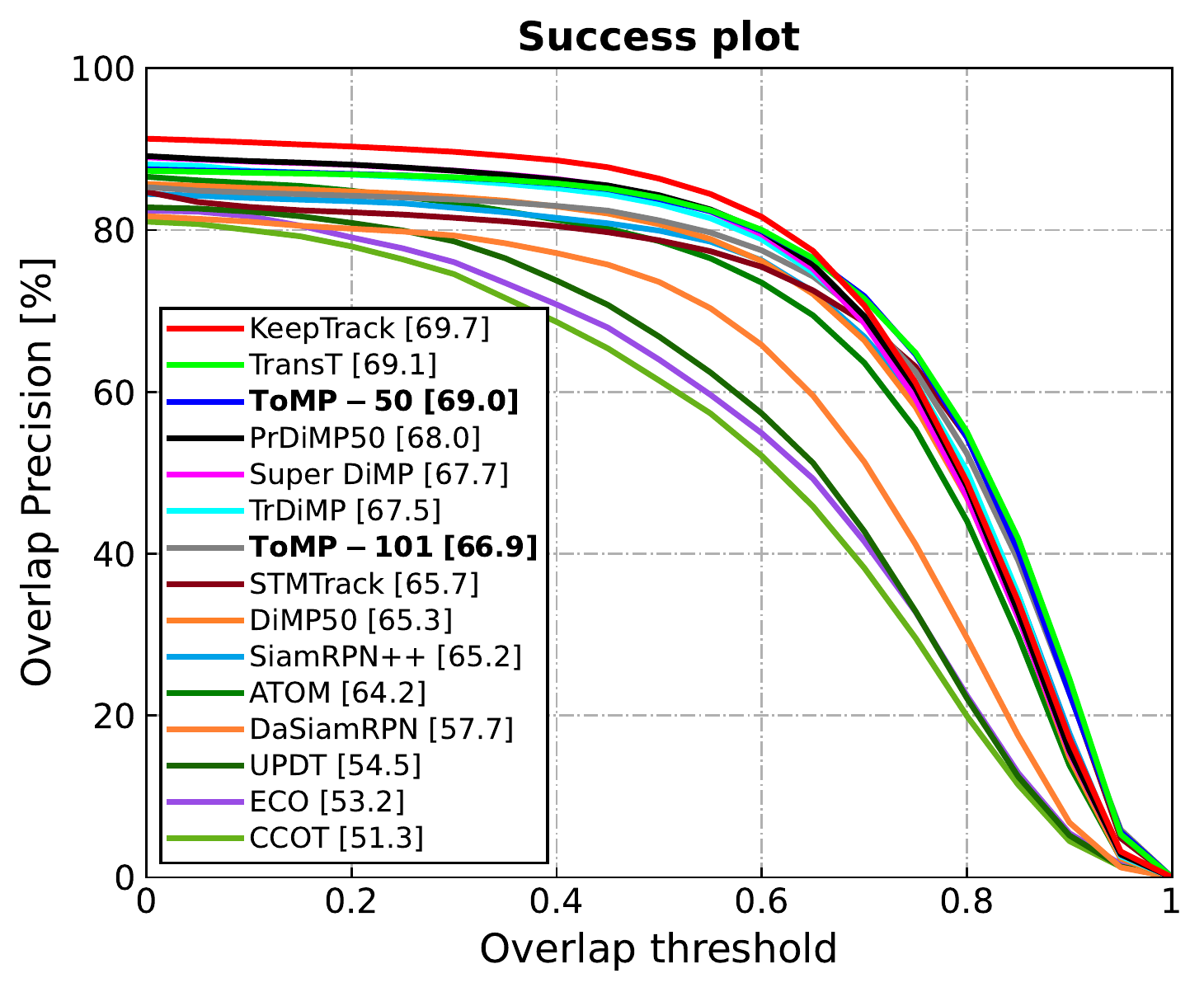}
    \caption{UAV123~\cite{Mueller_2016_ECCV_UAV123}}
    \label{sup:fig:uav-success}
  \end{subfigure}
  \begin{subfigure}{0.33\linewidth}
    \centering
    \includegraphics[width=\linewidth, keepaspectratio]{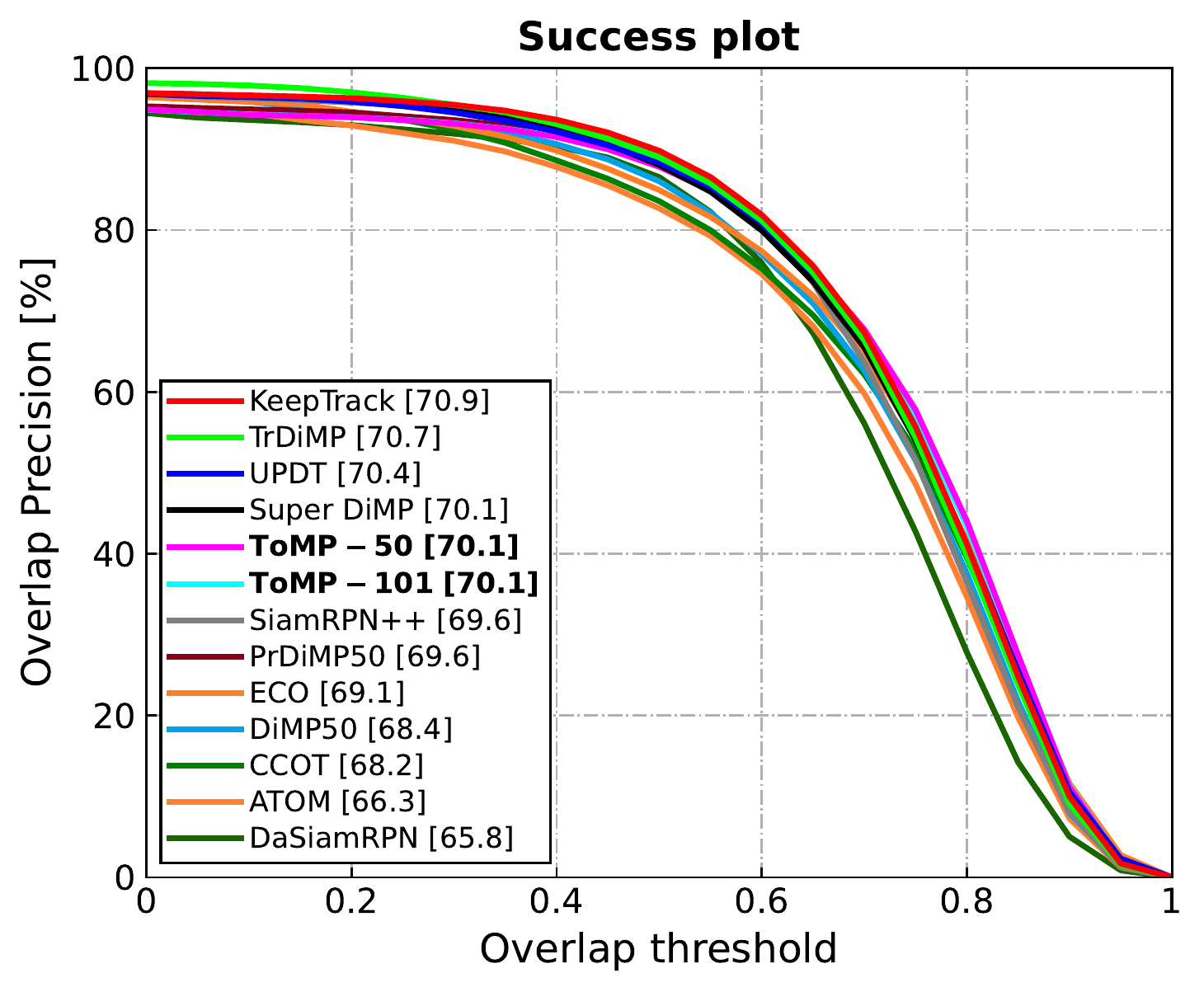}
    \caption{OTB-100~\cite{WU_2015_TPAMI_OTB}}
    \label{sup:fig:otb-success}
  \end{subfigure}
  \begin{subfigure}{0.33\linewidth}
    \centering
    \includegraphics[width=\linewidth, keepaspectratio]{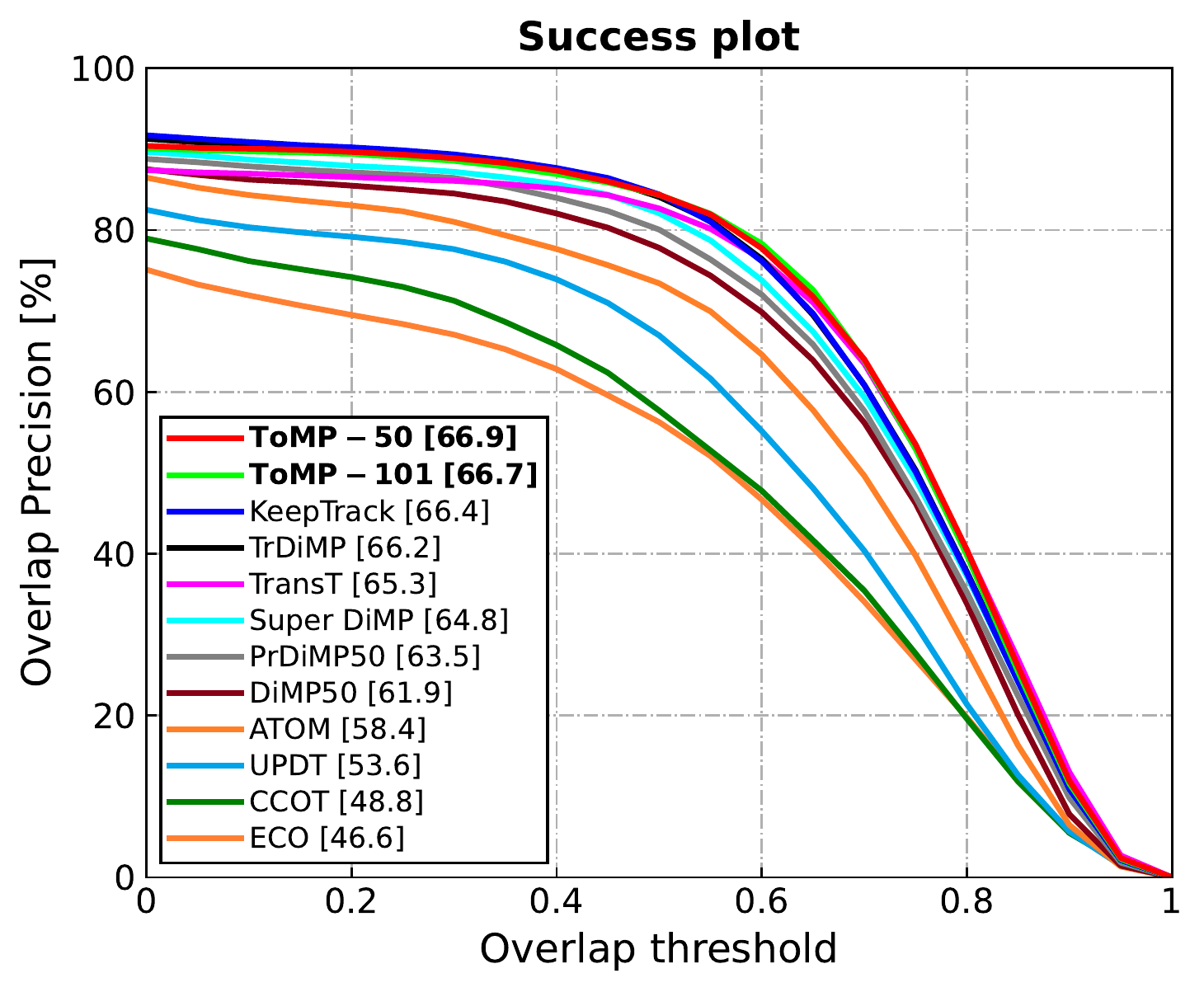}
    \caption{NFS~\cite{Galoogahi_2017_ICCV_NFS}}
    \label{sup:fig:nfs-success}
  \end{subfigure}
\caption{Success plots on the UAV123~\cite{Mueller_2016_ECCV_UAV123}, OTB-100~\cite{WU_2015_TPAMI_OTB} and NFS~\cite{Galoogahi_2017_ICCV_NFS} datasets in terms of overall AUC score, reported in the legend.}\label{sup:fig:uav_otb_nfs}
\end{figure*}

\subsection{Centerness}

Our proposed bounding box regression component is inspired by FCOS~\cite{Tian_2019_ICCV_FCOS} but in contrast to FCOS we omit an auxiliary centerness branch.  The classification head of FCOS is trained to predict a high score for almost every region inside the bounding box. The centerness branch is therefore needed to identify the center location of the object, used to select the bounding box offsets. In contrast, our classification branch is directly trained to accurately locate the object's center. The additional centerness branch is therefore redundant. Nonetheless, we train our best model with a centerness head and $L_\mathrm{centerness}$ and report the results in Tab.~\ref{sup:tab:centerness} (\nth{2}-\nth{4} rows).
The \nth{1} row shows the performance when omitting centerness for training.
We achieve comparable results when using the model trained with centerness but applying only the classification scores to localize the target (\nth{2} row). Using only the centerness scores decreases the performance (\nth{3} row) because centerness often fails to identify the target among distractors (see Fig.~\ref{sup:fig:centerness}). Finally, we follow FCOS and multiply the classification and centerness scores point-wise to retrieve the target object (\nth{4} row). We conclude that omitting the centerness branch for training and during inference to localize the target achieves the best tracking performance.

\section{Visual Results}\label{sup:sec:vis-results}
In this part we provide visual results of our tracker. First, we show three frames of different sequences where our tracker outperforms the state of the art. Secondly, we compare the produced target score map of our tracker with score maps obtained by optimization based model prediction. Finally, we show some failure cases of our tracker.

\begin{table}[!t]
	\centering
	\vspace{-0mm}
	\newcommand{\best}[1]{\textbf{\textcolor{red}{#1}}}
	\newcommand{\scnd}[1]{\textbf{\textcolor{blue}{#1}}}
	\newcommand{\dist}{\hspace{5pt}}%
	\resizebox{1.00\linewidth}{!}{%
        \begin{tabular}{l@{\dist}c@{\dist}c@{\dist}c@{\dist}c@{\dist}c@{\dist}c@{\dist}c@{\dist}c@{\dist}c@{\dist}c@{\dist}c@{\dist}c@{\dist}c@{\dist}c@{\dist}c@{\dist}c@{\dist}c@{\dist}c@{\dist}c@{\dist}c@{\dist}c@{\dist}c@{\dist}c@{\dist}c@{\dist}c@{\dist}c@{\dist}c@{\dist}c@{\dist}c@{\dist}}
        	\toprule
        	           & \textbf{ToMP}   & \textbf{ToMP}   &     & STARK        & STARK      & Ocean & Alpha  &      &      & Fast  \\
        	           & \textbf{101+AR} & \textbf{50 +AR} & RPT & ST50+AR      & ST101+AR   & Plus  & Refine & AFOD & LWTL & Ocean \\
        	           & & & \cite{Ma_2020_ECCVW_RPT,Kristan_2020_ECCVW_VOT2020} & \cite{Yan_2021_ICCV_STARK} & \cite{Yan_2021_ICCV_STARK} & \cite{Chen_2020_ECCVW_AFOD,Kristan_2020_ECCVW_VOT2020} & \cite{Yan_2021_CVPR_AlphaRefine,Kristan_2020_ECCVW_VOT2020} & \cite{Kristan_2020_ECCVW_VOT2020} & \cite{Bhat_2020_ECCV_LWL,Kristan_2020_ECCVW_VOT2020} & \cite{Kristan_2020_ECCVW_VOT2020} \\          
        	\midrule
        	EAO        & 0.497 & 0.496 & \best{0.530} & 0.505        & 0.497        & 0.491        & 0.482 & 0.472 & 0.463 & 0.461 \\
        	Accuracy   & 0.750 & 0.754 & 0.700        & \scnd{0.759} & \best{0.763} & 0.685        & 0.754 & 0.713 & 0.719 & 0.693 \\
        	Robustness & 0.798 & 0.793 & \best{0.869} & 0.817        & 0.789        & \scnd{0.842} & 0.777 & 0.795 & 0.798 & 0.803 \\
            \bottomrule
        \end{tabular}
	}
	\caption{Comparison to the state of the art of segmentation only methods on VOT2020ST~\cite{Kristan_2020_ECCVW_VOT2020} in terms of EAO score.
	}
	\label{sup:tab:vot2020}%
\end{table}
\begin{table*}[!t]
	\centering
	\newcommand{\best}[1]{\textbf{\textcolor{red}{#1}}}
	\newcommand{\scnd}[1]{\textbf{\textcolor{blue}{#1}}}
	\newcommand{\dist}{\hspace{5pt}}%
	\resizebox{1.00\textwidth}{!}{%
        \begin{tabular}{l@{\dist}c@{\dist}c@{\dist}c@{\dist}c@{\dist}c@{\dist}c@{\dist}c@{\dist}c@{\dist}c@{\dist}c@{\dist}c@{\dist}c@{\dist}c@{\dist}c@{\dist}c@{\dist}c@{\dist}c@{\dist}c@{\dist}c@{\dist}c@{\dist}c@{\dist}c@{\dist}c@{\dist}c@{\dist}c@{\dist}c@{\dist}c@{\dist}c@{\dist}c@{\dist}c@{\dist}c@{\dist}c@{\dist}c@{\dist}}
        	\toprule
        	        & \textbf{ToMP} & \textbf{ToMP} & Keep  & STARK & Tr   &        &      & STARK & Super       & Pr         & Siam  & STM   &      &      & Siam  &      &      & Retina      & FCOS \\
        	        & \textbf{101}  & \textbf{50}   & Track & ST101 & DiMP & TransT & SAOT & ST50  & DiMP        & DiMP       & R-CNN & Track & DiMP & KYS  & RPN++ & ATOM & UPDT & MAML        & MAML \\
        	        & & & \cite{Mayer_2021_ICCV_KeepTrack} & \cite{Yan_2021_ICCV_STARK} &\cite{Wang_2021_CVPR_TrDiMP} & \cite{Chen_2021_CVPR_TransT} & \cite{Zhou_2021_ICCV_SAOT} & \cite{Yan_2021_ICCV_STARK} &\cite{Danelljan_2019_github_pytracking} & \cite{Danelljan_2020_CVPR_PRDIMP} & \cite{Voigtlaender_2020_CVPR_SiamRCNN} & \cite{Fu_2021_CVPR_STMTrack} & \cite{Bhat_2019_ICCV_DIMP} & \cite{Bhat_2020_ECCV_KYS} & \cite{Li_2019_CVPR_SiamRPN++} & \cite{Danelljan_2019_CVPR_ATOM} & \cite{Bhat_2018_ECCV_UPDT} & \cite{Wang_2020_CVPR_MAML} & \cite{Wang_2020_CVPR_MAML} \\
        	\midrule
        	UAV123 & 66.9        & 69.0        & \best{69.7} & 68.2 & 67.5 & \scnd{69.1} & \scnd{69.1} & --   & 67.7 & 68.0 & 64.9 & 64.7        & 65.3 & --   & 61.3 & 64.2  & 54.5 & --   & --   \\
        	OTB-100& 70.1        & 70.1        & 70.9        & 68.1 & 71.1 & 69.4        & 68.5        & 71.4 & 70.1 & 69.6 & 70.1 & \scnd{71.9} & 68.4 & 69.5 & 69.6 & 66.9  & 70.2 & 71.2 & 70.4 \\
        	NFS    & \scnd{66.7} & \best{66.9} & 66.4        & 66.2 & 66.2 & 65.7        & 65.2        & 65.6 & 64.8 & 63.5 & 63.9 & --          & 62.0 & 63.5 & --   & 58.4  & 53.7 & --   & --   \\\bottomrule
        	\toprule
        	&       &       & Auto  & Auto  & Siam & Siam &      &       &        &             &     & Siam  &       &      & Siam & Siam &       &      & DaSiam \\
        	& Ocean & STN   & Match & Track & BAN  & CAR  & ECO  & DCFST & PG-NET & CRACT       & GCT & GAT   & CLNet & TLPG & AttN & FC++ & MDNet & CCOT & RPN    \\
        	& \cite{Zhang_2020_ECCV_Ocean} & \cite{Liu_2020_ECCV_STN} & \cite{Zhang_2021_ICCV_Match} & \cite{Li_2020_CVPR_AutoTrack} & \cite{Chen_2020_CVPR_SiamBAN} & \cite{Guo_2020_CVPR_SiamCAR} & \cite{Danelljan_2017_CVPR_ECO} & \cite{Zheng_2020_ECCV_DCFST} & \cite{Bingyan_2020_ECCV_PGNet} & \cite{Fan_2020_arxiv_CRACT} & \cite{Gao_2019_CVPR_GCT} & \cite{Guo_2020_arxiv_SiamGAT} & \cite{Dong_2020_ECCV_CLNet} & \cite{Li_2020_IJCAI_TLPG} & \cite{Yu_2020_CVPR_SiamAttN} & \cite{Xu_2020_AAAI_SiamFCpp} & \cite{Nam_2016_CVPR_MDNet} & \cite{Danelljan_2016_ECCV_CCOT} & \cite{Zhu_2018_ECCV_DaSiamRPN} \\
        	\midrule
        	UAV123 & --   & 64.9  & --   & 67.1 & 63.1 & 61.4 & 53.2 & --   & --   & 66.4        & 50.8 & 64.6 & 63.3 & --   & 65.0 & --   & --   & 51.3 & 57.7 \\
        	OTB-100& 68.4 & 69.3  & 71.4 & --   & 69.6 & --   & 69.1 & 70.9 & 69.1 & \best{72.6} & 64.8 & 71.0 & --   & 69.8 & 71.2 & 68.3 & 67.8 & 68.2 & 65.8 \\
        	NFS    & --   & --    & --   & --   & 59.4 & --   & 46.6 & 64.1 & --   & 62.5        & --   & --   & 54.3 & --   & --   & --   & 41.9 & 48.8 & --   \\
        	\bottomrule
        \end{tabular}
	}
	\caption{Comparison with state-of-the-art on the OTB-100~\cite{WU_2015_TPAMI_OTB}, NFS~\cite{Galoogahi_2017_ICCV_NFS} and UAV123~\cite{Mueller_2016_ECCV_UAV123} datasets in terms of overall AUC score.}
	\label{sup:tab:otb_nfs_uav}%
\end{table*}

\begin{table*}[!t]
	\centering
	\newcommand{\best}[1]{\textbf{\textcolor{red}{#1}}}
	\newcommand{\scnd}[1]{\textbf{\textcolor{blue}{#1}}}
	\newcommand{\dist}{\hspace{5pt}}%
	\resizebox{1.00\textwidth}{!}{%
        \begin{tabular}{l@{\dist}c@{\dist}c@{\dist}c@{\dist}c@{\dist}c@{\dist}c@{\dist}c@{\dist}c@{\dist}c@{\dist}c@{\dist}c@{\dist}c@{\dist}c@{\dist}c@{\dist}c@{\dist}c@{\dist}c@{\dist}c@{\dist}c@{\dist}c@{\dist}c@{\dist}c@{\dist}c@{\dist}c@{\dist}c@{\dist}c@{\dist}c@{\dist}c@{\dist}c@{\dist}c@{\dist}c@{\dist}c@{\dist}c@{\dist}c@{\dist}c@{\dist}}
        	\toprule
        	        & \textbf{ToMP} & \textbf{ToMP} & STARK & Keep  & STARK & Alpha  &        & Siam  & Tr   & Super &      & STM   &     & Pr   & DM    & Auto  &      &      &      \\ 
        	        & \textbf{101}  & \textbf{50}   & ST101 & Track & ST50  & Refine & TransT & R-CNN & DiMP & Dimp  & SAOT & Track & DTT & DiMP & Track & Match & TLPG & TACT & LTMU \\
        	        &               &               & \cite{Yan_2021_ICCV_STARK} & \cite{Mayer_2021_ICCV_KeepTrack} & \cite{Yan_2021_ICCV_STARK} & \cite{Yan_2021_CVPR_AlphaRefine} & \cite{Chen_2021_CVPR_TransT} & \cite{Voigtlaender_2020_CVPR_SiamRCNN} & \cite{Wang_2021_CVPR_TrDiMP} & \cite{Danelljan_2019_github_pytracking} & \cite{Zhou_2021_ICCV_SAOT} & \cite{Fu_2021_CVPR_STMTrack} & \cite{Yu_2021_ICCV_HPF} & \cite{Danelljan_2020_CVPR_PRDIMP} & \cite{Zhang_2021_CVPR_DMTrack} & \cite{Zhang_2021_ICCV_Match} &\cite{Li_2020_IJCAI_TLPG} & \cite{Choi_2020_ACCV_TACT} & \cite{Dai_2020_CVPR_LTMU} \\
        	\midrule
        	LaSOT   & \best{68.5} & \scnd{67.6} & 67.1 & 67.1 & 66.4 &  65.3 & 64.9 & 64.8 & 63.9 & 63.1 & 61.6 & 60.6 & 60.1 & 59.8 & 58.4 & 58.3 & 58.1 & 57.5 & 57.2  \\ \bottomrule
        	\toprule
        	&      &       & Siam &       & Siam & Siam  & PG  & FCOS & Global &      & DaSiam & Siam & Siam &       & Siam   & Retina & Siam &        &     \\
        	& DiMP & Ocean & AttN & CRACT & FC++ & GAT   & NET & MAML & Track  & ATOM & RPN    & BAN  & CAR  & CLNet & RPN++  & MAML   & Mask & ROAM++ & SPLT\\
        	& \cite{Bhat_2019_ICCV_DIMP} & \cite{Zhang_2020_ECCV_Ocean} & \cite{Yu_2020_CVPR_SiamAttN} & \cite{Fan_2020_arxiv_CRACT} & \cite{Xu_2020_AAAI_SiamFCpp} & \cite{Guo_2020_arxiv_SiamGAT} & \cite{Bingyan_2020_ECCV_PGNet} & \cite{Wang_2020_CVPR_MAML} & \cite{Huang_2020_AAAI_GlobalTrack} & \cite{Danelljan_2019_CVPR_ATOM} & \cite{Zhu_2018_ECCV_DaSiamRPN}$^\dagger$ & \cite{Chen_2020_CVPR_SiamBAN} & \cite{Guo_2020_CVPR_SiamCAR} & \cite{Dong_2020_ECCV_CLNet} & \cite{Li_2019_CVPR_SiamRPN++}$^\dagger$ & \cite{Wang_2020_CVPR_MAML} & \cite{Wang_2019_CVPR_SiamMask}$^\dagger$ & \cite{Yang_2020_CVPR_ROAM} & \cite{Yan_2019_ICCV_SPLT}\\
        	\midrule
        	LaSOT  & 56.9 & 56.0  & 56.0 & 54.9  & 54.4 & 53.9  & 53.1   & 52.3 & 52.1   & 51.5 & 51.5 & 51.4 & 50.7 & 49.9  & 49.6   & 48.0   & 46.7 & 44.7 & 42.6\\\bottomrule
        \end{tabular}
	}
	\caption{Comparison with state-of-the-art on the LaSOT~\cite{Fan_2019_CVPR_Lasot} test set in terms of overall AUC score. The symbol $^\dagger$ marks results that were produced by Fan~\etal~\cite{Fan_2019_CVPR_Lasot} otherwise they are obtained directly from the official paper.}
	\label{sup:tab:lasot}%
	\vspace{-0mm}
\end{table*}
\subsection{Visual Comparison to the State of the Art}\label{sup:sec:vis-comp}

Fig.~\ref{sup:fig:visual-results} shows three frames of eight different LaSOT~\cite{Fan_2019_CVPR_Lasot} sequences where each frame contains the ground truth annotation of the target object and the predictions of three different trackers: SuperDiMP~\cite{Danelljan_2019_github_pytracking}, STARK-ST101~\cite{Yan_2021_ICCV_STARK} and ToMP-101. We observe that our tracker produces in most sequences more robust and in some more accurate bounding box predictions than the related methods. In particular it achieves solid robustness for scenarios where distractors are present but the target object is at least partially visible and not undergoing a full occlusion.  

\begin{figure*}[t]
\centering
  \begin{subfigure}{0.49\linewidth}
    \centering
    \includegraphics[width=\linewidth, keepaspectratio]{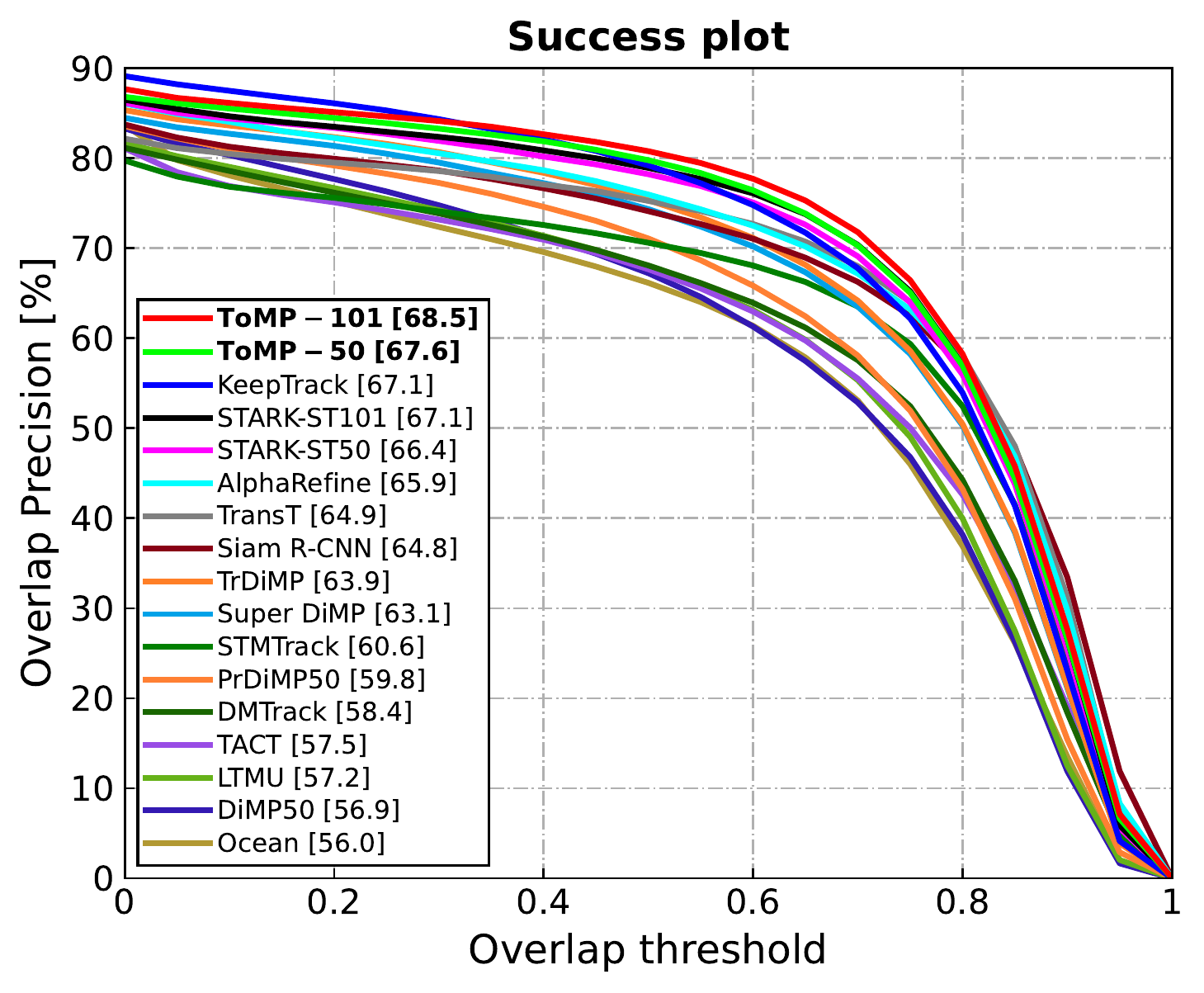}
    \caption{Success}
    \label{sup:fig:lasot-success}
  \end{subfigure}
  \begin{subfigure}{0.49\linewidth}
    \centering
    \includegraphics[width=\linewidth, keepaspectratio]{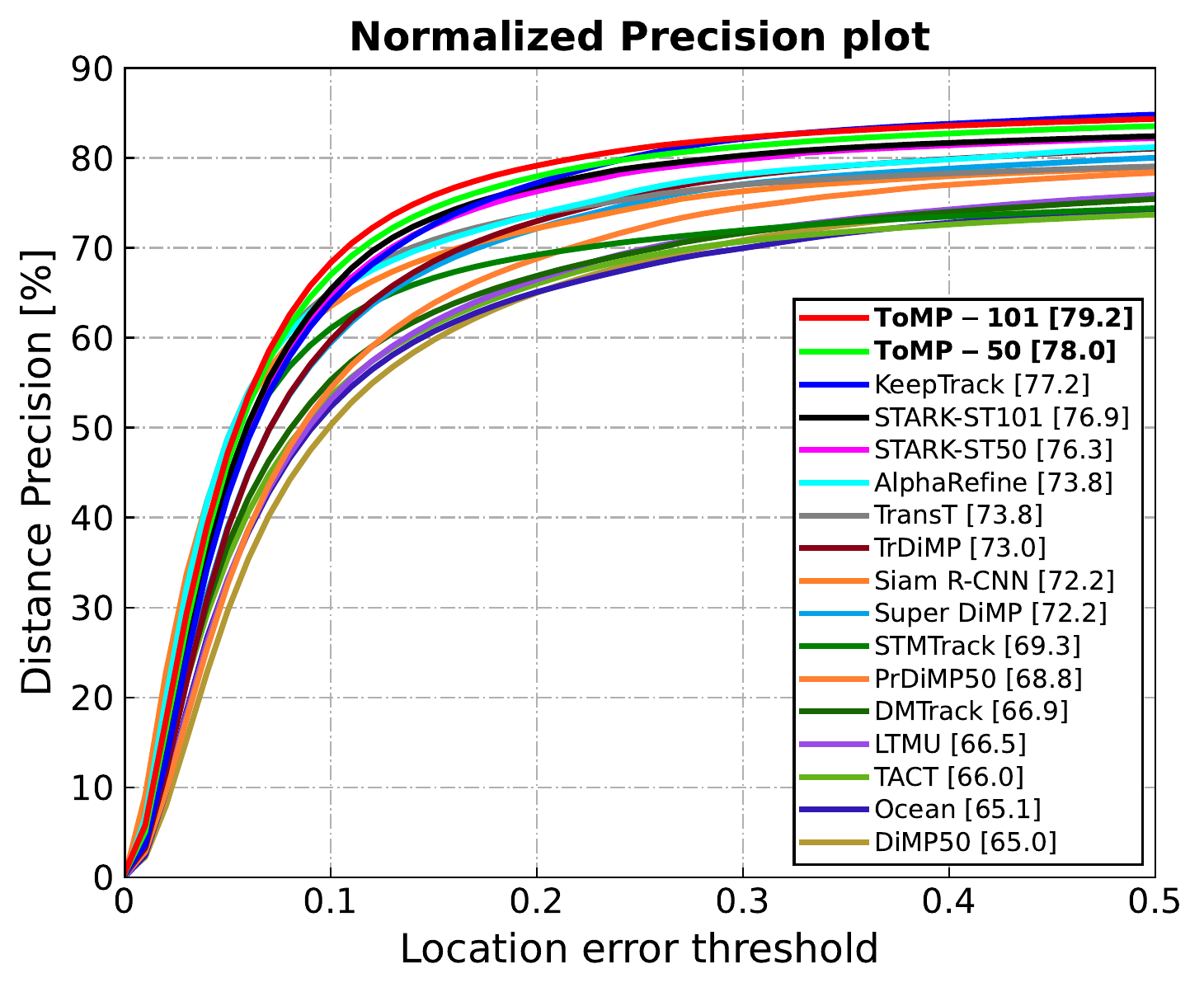}
    \caption{Normalized Precision}
    \label{sup:fig:lasot-norm-prec}
  \end{subfigure}
\caption{Success and normalized precision plots on LaSOT~\cite{Fan_2019_CVPR_Lasot}. Our approach outperforms all other methods by a large margin in AUC, reported in the legend.}\label{sup:fig:lasot}
\vspace{-5mm}
\end{figure*}

\subsection{Target Model Prediction}\label{sup:sec:score-map}
Fig.~\ref{sup:fig:score_maps} shows the target score maps produced by the target model when using two different model predictors for three different sequences. In detail we compare the target score map produced by SuperDiMP~\cite{Danelljan_2019_github_pytracking} that adopts the DiMP~\cite{Bhat_2019_ICCV_DIMP} model predictor with optimized settings. In particular it uses a slightly smaller search area factor of 6 instead of 5 and a target score resolution of 22 instead of 18. Note, that our tracker uses 5 and 18 similar to DiMP~\cite{Bhat_2019_ICCV_DIMP} as stated Sec.~\ref{sup:sec:inf-details}. We observe that our model predictor leads to much cleaner and unambiguous target localization than DiMP. While the former often produces multiple local maxima for distractors, our methods is able to almost fully suppress these. An important design choice that enables this is the transductive model weight and test feature prediction produced by our Transformer based model predictor. However, the cleaner score maps come with the risk, that once the target is lost and a distractor is tracked instead recovering is less likely since our tracker effectively suppresses distractors. Similarly, our method learns to produce a score map containing a Gaussian such that overall the maximum score values are higher than by SuperDiMP. Thus, we chose a relatively high threshold to decide whether to use a previous prediction as training sample or not.   
\subsection{Failure Cases}\label{sup:sec:failure-cases}

Fig.~\ref{sup:fig:failure_cases} shows failure cases of our tracker. In particular, it shows three frames of four different LaSOT~\cite{Fan_2019_CVPR_Lasot}
sequences containing the ground truth annotations and the predicted bounding boxes of our tracker using a ResNet-101~\cite{He_2016_CVPR_Resnet} as backbone. To summarize, our tracker typically fails if object similar to the targets so called distractors are present. While the sole presence of distractors typically does not lead to tracking failure, our tracker shows difficulties in sequences where the target is occluded and distractors are present (\nth{1} and \nth{3} row). Instead of detecting that the target is occluded the tracker starts to track a distractor instead.
Another challenging scenario are sequences where the target and a distractor approach each other (\nth{2} row in Fig.~\ref{sup:fig:failure_cases}) or one occludes the other (\nth{4} row in Fig.~\ref{sup:fig:failure_cases}). The model then detects only a single object instead of two in both scenarios. Once they diverge again and the tracker detects two objects it typically fails to reliably differentiate between the target and the distractor.

\section{Experiments}\label{sup:sec:exps}
We provide more detailed experiments to complement the comparison to the state of-the art performed in the main paper. And provide results
for the VOT2020ST~\cite{Kristan_2020_ECCVW_VOT2020} challenge when using AlphaRefine~\cite{Yan_2021_CVPR_AlphaRefine} on top of our method in order to compare with methods that produce a segmentation mask as output.

\subsection{VOT2020 with AlphaRefine}\label{sup:sec:vot2020}
\begin{figure*}[!t]
\centering
  \begin{subfigure}{0.49\linewidth}
    \centering
    \includegraphics[width=\linewidth, keepaspectratio]{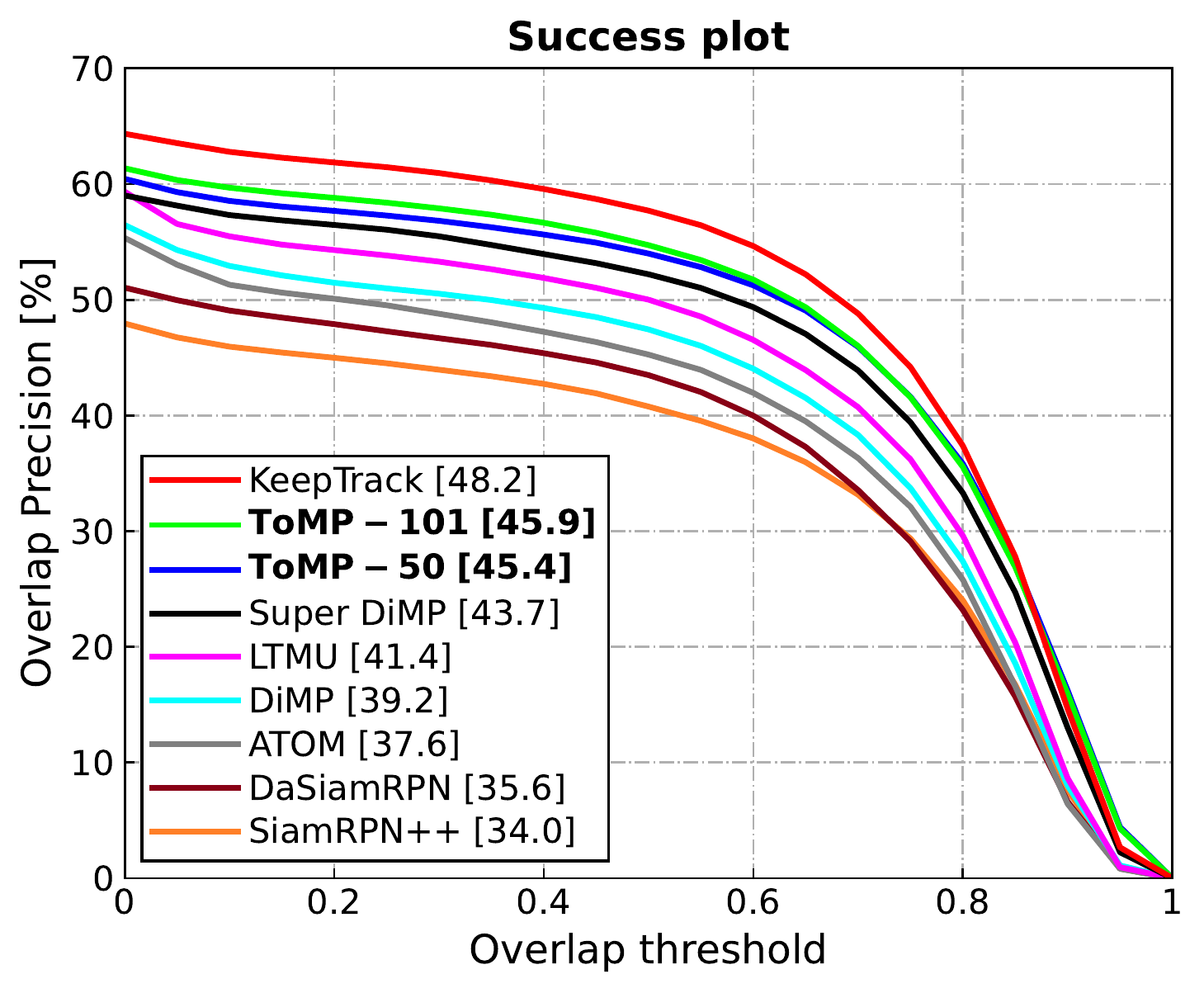}
    \caption{Success}
    \label{sup:fig:lasotextsub-success}
  \end{subfigure}
  \begin{subfigure}{0.49\linewidth}
    \centering
    \includegraphics[width=\linewidth, keepaspectratio]{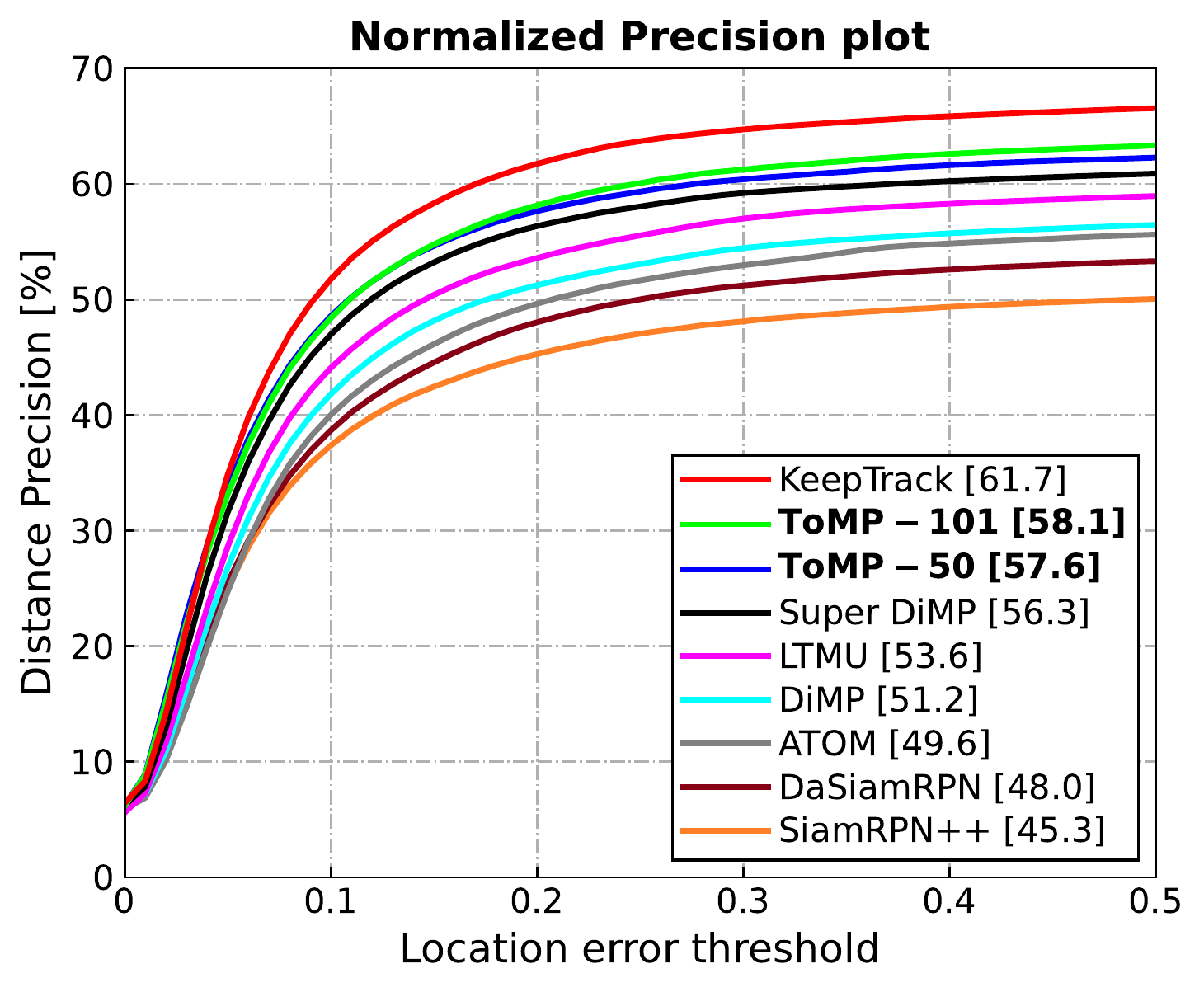}
    \caption{Normalized Precision}
    \label{sup:fig:lasotextsub-norm-prec}
  \end{subfigure}
\caption{Success and normalized precision plots on LaSOTExtSub~\cite{Fan_2020_IJCV_Lasot_ext}. Our approach outperforms all other methods by a large margin in AUC, reported in the legend.}\label{sup:fig:lasotextsub}
\end{figure*}

\begin{table*}[t]
	\centering
	\newcommand{\best}[1]{\textbf{\textcolor{red}{#1}}}
	\newcommand{\scnd}[1]{\textbf{\textcolor{blue}{#1}}}
	\newcommand{\dist}{\hspace{4pt}}%
	\resizebox{1.00\textwidth}{!}{%
        \begin{tabular}{l@{\dist}c@{\dist}c@{\dist}c@{\dist}c@{\dist}c@{\dist}c@{\dist}c@{\dist}c@{\dist}c@{\dist}c@{\dist}c@{\dist}c@{\dist}c@{\dist}c@{\dist}|c@{\dist}}
        	\toprule
        	                       & Illumination & Partial     &                & Motion         & Camera      &             & Background & Viewpoint   & Scale        & Full        & Fast        &             & Low         & Aspect        &       \\
        	                       & Variation    & Occlusion   & Deformation    & Blur           & Motion      & Rotation    & Clutter    & Change      & Variation    & Occlusion   & Motion      & Out-of-View & Resolution  & Ration Change & Total \\
 
        	\midrule
            LTMU                   & 56.5         & 54.0        & 57.2           & 55.8           & 61.6        & 55.1        & 49.9        & 56.7        & 57.1        & 49.9        & 44.0        & 52.7        & 51.4        & 55.1          & 57.2 \\
            PrDiMP50               & 63.7         & 56.9        & 60.8           & 57.9           & 64.2        & 58.1        & 54.3        & 59.2        & 59.4        & 51.3        & 48.4        & 55.3        & 53.5        & 58.6          & 59.8 \\
            STMTrack               & 65.2         & 57.1        & 64.0           & 55.3           & 63.3        & 60.1        & 54.1        & 58.2        & 60.6        & 47.8        & 42.4        & 51.9        & 50.3        & 58.8          & 60.6 \\
            SuperDiMP              & 67.8         & 59.7        & 63.4           & 62.0           & 68.0        & 61.4        & 57.3        & 63.4        & 62.9        & 54.1        & 50.7        & 59.0        & 56.4        & 61.6          & 63.1 \\
            TrDiMP                 & 67.5         & 61.1        & 64.4           & 62.4           & 68.1        & 62.4        & 58.9        & 62.8        & 63.4        & 56.4        & 53.0        & 60.7        & 58.1        & 62.3          & 63.9 \\
            Siam R-CNN             & 64.6         & 62.2        & 65.2           & 63.1           & 68.2        & 64.1        & 54.2        & 65.3        & 64.5        & 55.3        & 51.5        & 62.2        & 57.1        & 63.4          & 64.8 \\
            TransT                 & 65.2         & 62.0        & 67.0           & 63.0           & 67.2        & 64.3        & 57.9        & 61.7        & 64.6        & 55.3        & 51.0        & 58.2        & 56.4        & 63.2          & 64.9 \\
            AlphaRefine            & \scnd{69.4}  & 62.3        & 66.3           & \scnd{65.2}    & 70.0        & 63.9        & 58.8        & 63.1        & 65.4        & 57.4        & 53.6        & 61.1        & 58.6        & 64.1          & 65.3 \\
            STARK-ST50             & 66.8         & 64.3        & 66.9           & 62.9           & 69.0        & 66.1        & 57.3        & 67.8        & 66.1        & 58.7        & 53.8        & 62.1        & 59.4        & 64.9          & 66.4 \\
            STARK-ST101            & 67.5         & \scnd{65.1} & 68.3           & 64.5           & 69.5        & 66.6        & 57.4        & \scnd{68.8} & 66.8        & 58.9        & 54.2        & 63.3        & 59.6        & 65.6          & 67.1 \\
            KeepTrack              & \best{69.7}  & 64.1        & 67.0           & \best{66.7}    & \scnd{71.0} & 65.3        & \scnd{61.2} & 66.9        & 66.8        & \best{60.1} & \scnd{57.7} & \best{64.1} & \scnd{62.0} & 65.9          & 67.1 \\
            \textbf{ToMP-50}       & 66.8         & 64.9        & \scnd{68.5}    & 64.6           & 70.2        & \scnd{67.3} & 59.1        & 67.2        & \scnd{67.5} & \scnd{59.3} & 56.1        & \scnd{63.7} & 61.1        & \scnd{66.5}   & \scnd{67.6} \\
            \textbf{ToMP-101}      & 69.0         & \best{65.3} & \best{69.4}    & \scnd{65.2}    & \best{71.7} & \best{67.8} & \best{61.5} & \best{69.2} & \best{68.4} & 59.1        & \best{57.9} & \best{64.1} & \best{62.5} & \best{67.2}   & \best{68.5} \\

            \bottomrule
        \end{tabular}
	}
	\caption{LaSOT~\cite{Fan_2019_CVPR_Lasot} attribute-based analysis. Each column corresponds to the results computed on all sequences in the dataset with the corresponding attribute.}
	\label{sup:tab:lasot_attributes}%
\end{table*}
In contrast to previous years where the sequences in the VOT short-term challenge were annotated with bounding boxes~\cite{Matej_2018_ECCVW_VOT2018,Kristan_2019_ICCVW_VOT2019} the sequences of the more recent challenges contain segmentation mask annotations~\cite{Kristan_2020_ECCVW_VOT2020,Kristan_2021_ICCVW_VOT2021} of the target in each frame. In the main paper we compare our method with methods that produce bounding boxes. Thus, in addition, we compare our method on the VOT2020 short-term challenge to methods that produce a segmentation mask in each frame.
Since our method produces only a bounding box, we use AlphaRefine~\cite{Yan_2021_CVPR_AlphaRefine} that is able to produce a segmentation mask give the bounding box. Tab.~\ref{sup:tab:vot2020} shows that our method achieves competitive results. In particular ToMP-101 achieves the same EAO (for more details on EAO we refer the reader to~\cite{Kristan_2020_ECCVW_VOT2020}) as STARK-ST101+AR~\cite{Yan_2021_ICCV_STARK} that employs AlphaRefine too. Nonetheless, RPT~\cite{Ma_2020_ECCVW_RPT} achieves higher EAO than our tracker. In particular it scores a higher robustness but a lower accuracy than our trackers.   

\subsection{UAV123, OTB-100 and NFS}\label{sup:sec:uav-otb-nfs}

To complement the results detailed in the paper, we provide the success plots for the UAV123~\cite{Mueller_2016_ECCV_UAV123} dataset in Fig.~\ref{sup:fig:uav-success}, the OTB-100~\cite{WU_2015_TPAMI_OTB} dataset in Fig.~\ref{sup:fig:otb-success} and the NFS~\cite{Galoogahi_2017_ICCV_NFS} dataset in Fig.~\ref{sup:fig:nfs-success}. Fig.~\ref{sup:fig:uav-success} shows that KeepTrack~\cite{Mayer_2021_ICCV_KeepTrack} and PrDiMP50~\cite{Danelljan_2020_CVPR_PRDIMP} achieve higher robustness than our tracker ($T < 0.6$) but that our trackers together with TransT~\cite{Chen_2021_CVPR_TransT} reaches the highest accuracy among all trackers ($T > 0.7$) compensating for the lower robustness. Fig.~\ref{sup:fig:otb-success} reveals similar conclusions on OTB-100. For NFS Fig.~\ref{sup:fig:nfs-success} shows that our tracker is almost as robust as KeepTrack~\cite{Mayer_2021_ICCV_KeepTrack} but achieves superior accuracy leading to a new state of the art. While we reported only the methods with the highest performances on these datasets in the main paper, we compare our method in Tab.~\ref{sup:tab:otb_nfs_uav} with additional related methods.

\subsection{LaSOT and LaSOTExtSub}\label{sup:sec:lasot}

In addition to the success plots, we provide the normalized precision plots on the LaSOT~\cite{Fan_2019_CVPR_Lasot} test set in Fig.~\ref{sup:fig:lasot} the LaSOTExtSub~\cite{Fan_2020_IJCV_Lasot_ext} test set in Fig.~\ref{sup:fig:lasotextsub}. The normalized precision score $\mathrm{NPr}_D$ measures the percentage of frames where the normalized distance (relative to the target size) between the predicted and ground-truth target center location is less than a threshold $D\in [0, 0.5]$. The ranking is determined by computing the AUC of each tracker.
The AUC is reported in the legend of Figs.~\ref{sup:fig:lasot-norm-prec} and~\ref{sup:fig:lasotextsub-norm-prec}. We compare our tracker on LaSOT with the state of the art in Tab.~\ref{sup:tab:lasot} and show their performance if available in Fig.~\ref{sup:fig:lasot}.
In Fig.~\ref{sup:fig:lasotextsub} we show results of methods produced by Fan~\etal~\cite{Fan_2020_IJCV_Lasot_ext} except KeepTrack~\cite{Mayer_2021_ICCV_KeepTrack} and SuperDiMP~\cite{Danelljan_2019_github_pytracking} that we obtained from Mayer~\etal~\cite{Mayer_2021_ICCV_KeepTrack}.

\subsubsection{Attributes}\label{sup:sec:lasot-attributes}

To support the attribute based analysis in the main paper, where we compared the performance of our tracker with other
Transformer based trackers, we provide the detailed analysis for multiple trackers and ToMP in Tab.~\ref{sup:tab:lasot_attributes}.
ToMP-101 achieves the best performance on all but three. It achieves the second best results for \textit{Motion Blur} behind KeepTrack~\cite{Mayer_2021_ICCV_KeepTrack} and similar to AlphaRefine~\cite{Yan_2021_CVPR_AlphaRefine}. Further ToMP-101 achieves the third best for \textit{Full Occlusion} behind KeepTrack~\cite{Mayer_2021_ICCV_KeepTrack} and ToMP-50. Similarly it scores third for \textit{Illumination Variation} behind KeepTrack~\cite{Mayer_2021_ICCV_KeepTrack} and AlphaRefine~\cite{Yan_2021_CVPR_AlphaRefine}. We further observe, that discriminative model prediction based methods such as TrDiMP~\cite{Wang_2021_CVPR_TrDiMP}, SuperDiMP~\cite{Danelljan_2019_github_pytracking}, AlphaRefine~\cite{Yan_2021_CVPR_AlphaRefine}, KeepTrack~\cite{Mayer_2021_ICCV_KeepTrack} and ToMP all outperform STARK~\cite{Yan_2021_ICCV_STARK} on the attribute \textit{Background Clutter} showing the advantage of using full training samples during tracking instead of cropped templates that mainly cover the centered target.
\end{appendices}

\end{document}